\documentclass[10pt, logo, twocolumn, copyright]{nvidiatechreport}

\usepackage{mdframed}
\usepackage[utf8]{inputenc} 
\usepackage[T1]{fontenc}    

\usepackage{amsfonts}       
\usepackage{nicefrac}       
\usepackage{microtype}      
\usepackage[dvipsnames]{xcolor}         
\usepackage{multirow}
\usepackage{multicol}
\usepackage{graphicx}
\usepackage[numbers]{natbib}
\usepackage{tabto}
\usepackage{xspace}
\usepackage{amsmath}
\usepackage{adjustbox}
\usepackage{enumitem}
\usepackage{wrapfig}
\usepackage{dblfloatfix}
\usepackage{makecell}
\usepackage{subcaption}

\usepackage{footmisc}
\usepackage{tcolorbox}
\usepackage{enumitem}
\usepackage{float}

\usepackage[ruled,vlined,linesnumbered]{algorithm2e}

\usepackage{natbib}

\usepackage{amsmath,amsfonts,bm}

\def\eqref#1{equation~\ref{#1}}

\def\1{\bm{1}}

\DeclareMathAlphabet{\mathsfit}{\encodingdefault}{\sfdefault}{m}{sl}
\SetMathAlphabet{\mathsfit}{bold}{\encodingdefault}{\sfdefault}{bx}{n}

\usepackage{hyperref}
\usepackage{url}
\usepackage{booktabs}
\usepackage{multirow}
\usepackage{enumitem}
\usepackage{adjustbox}
\usepackage{tabularx}
\usepackage{arydshln}
\usepackage{wrapfig}

\usepackage{multirow}
\usepackage{colortbl}
\usepackage{amsmath}
\usepackage{makecell}
\usepackage{hhline}
\usepackage{array}
\usepackage{diagbox}
\usepackage{graphicx}
\usepackage{adjustbox}

\usepackage{colortbl} 
\usepackage[table]{xcolor} 

\usepackage[flushleft]{threeparttable} 
\usepackage{multirow}

\usepackage{wrapfig}

\definecolor{Note_color}{rgb}{0.0, 0.0, 0.0}

\definecolor{lightyellow}{RGB}{255, 255, 204}
\definecolor{lightred}{RGB}{255, 204, 204}
\definecolor{lightgreen}{RGB}{204, 255, 204}

\usepackage{xcolor}
\definecolor{maskorange}{HTML}{ffe6cc}
\definecolor{maskblue}{HTML}{d4e1f5}
\definecolor{maskred}{HTML}{ffcccc}
\definecolor{maskgreen}{HTML}{d5e7d4}
\definecolor{diff_green}{HTML}{7cb342}
\definecolor{nldgreen}{HTML}{E8F5E9}

\newcommand{\tinybox}[2][1.5ex]{%
  \begingroup
  \setlength{\fboxsep}{0pt}%
  \colorbox{#2}{\rule{0pt}{#1}\rule{#1}{0pt}}%
  \endgroup
}

\usepackage{pifont}
\newcommand\mynuma[1]{\ifcase#1 \or \ding{172}\or \ding{173}\or
  \ding{174}\or \ding{175}\or \ding{176}\or \ding{177}
  \or \ding{178}\or \ding{179}\or \ding{180}\or \ding{181}\else *\fi\relax}

\newcommand\mynumb[1]{\ifcase#1 \or \ding{182}\or \ding{183}\or
  \ding{184}\or \ding{185}\or \ding{186}\or \ding{187}
  \or \ding{188}\or \ding{189}\or \ding{190}\or \ding{191}\else *\fi\relax}

\newcommand{\METHOD}{Nemotron-Labs-Diffusion}

\title{\METHOD{}: A Tri-Mode Language Model Unifying Autoregressive, Diffusion, and Self-Speculation Decoding}

\author{Yonggan Fu, Lexington Whalen$^{1\dagger}$, Abhinav Garg, Chengyue Wu$^{2\dagger}$, Maksim Khadkevich, Nicolai Oswald, Enze Xie, Daniel Egert, Sharath Turuvekere Sreenivas, Shizhe Diao, Chenhan Yu, Ye Yu, Weijia Chen, Sajad Norouzi, Jingyu Liu$^{3\dagger}$, Shiyi Lan, Ligeng Zhu, Jin Wang$^{2\dagger}$, Jindong Jiang, Morteza Mardani, Mehran Maghoumi, Song Han$^4$, Ante Jukić, Nima Tajbakhsh, Jan Kautz, Pavlo Molchanov}

\correspondingauthor{X}

\begin{abstract}

We introduce \METHOD{}, a tri-mode language model (LM) that unifies AR, diffusion, and self-speculation decoding within a single architecture. Trained with a joint AR-diffusion objective, \METHOD{} can switch modes to sustain high throughput across deployment settings and concurrency levels. Our study shows that (1) AR and diffusion objectives are complementary: diffusion improves lookahead planning, while AR provides left-to-right linguistic priors. (2) In self-speculation mode, diffusion drafts while AR verifies, outperforming multi-token prediction (MTP) methods in both acceptance rate and real-device efficiency. (3) A speed-of-light analysis further demonstrates diffusion’s long-term potential, with up to 76.5\% more tokens per forward pass than self-speculation under an optimal sampler. Scaling to 3B, 8B, and 14B parameters, our \METHOD{} family, including base, instruct, and vision-language models, consistently outperforms state-of-the-art open-source AR and diffusion LMs in both accuracy and speed. For example, \METHOD{}-8B decodes $6\times$ more tokens per forward than Qwen3-8B with comparable accuracy, translating to $4\times$ higher throughput on SPEED-Bench with SGLang on a GB200 GPU.

\vspace{2mm}
\textbf{Models on Hugging Face:} \href{https://huggingface.co/collections/nvidia/nemotron-labs-diffusion}{\METHOD{} Model Family}

\end{abstract}

\newcommand{\teaserfigure}{%
  \begingroup
  \captionsetup{type=figure}%
  \centering
  \vspace{-1em}
  \includegraphics[width=\linewidth]{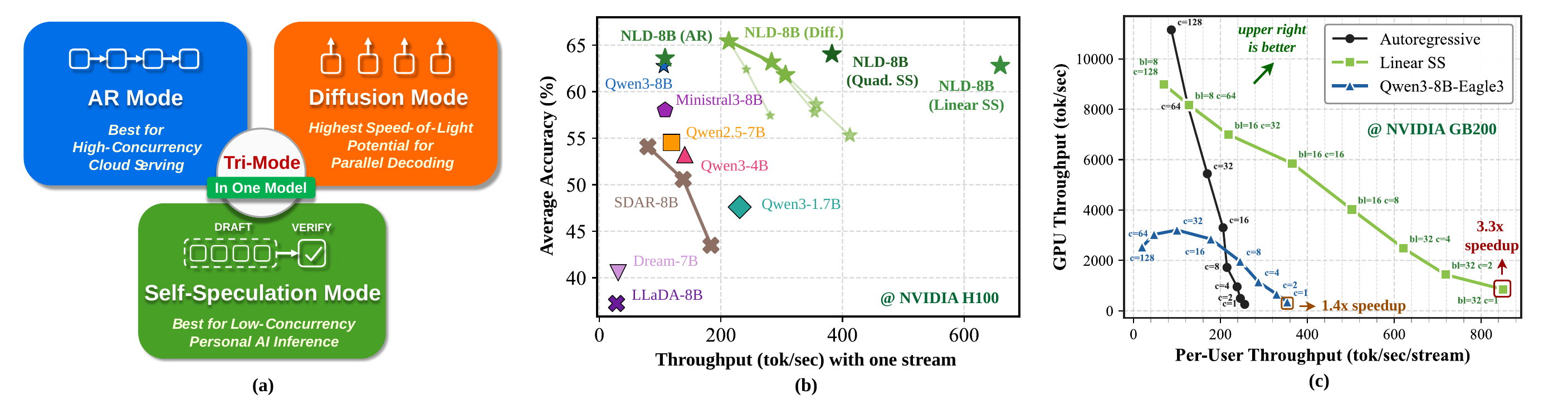}%
  \vspace{-0.5em}
  \captionof{figure}{%
    \textbf{(a)} An illustration of three modes in one model.
    \textbf{(b)} Accuracy--throughput trade-off measured on general benchmarks at batch size~1 on an NVIDIA H100 using PyTorch. NLD denotes our model and Linear/Quad SS denotes linear/quadratic self-speculation. \textcolor{diff_green}{$\bigstar$} indicates the diffusion mode with different denoising block sizes (8, 16, 32), where smaller symbols correspond to smaller block sizes.
    \textbf{(c)} Trade-off between system vs. per-user throughput of the AR and Linear SS modes of our 8B model and Qwen3-8B Eagle3, measured on SPEED-Bench~\cite{abramovich2026speed} at different concurrency $c$ on an NVIDIA GB200 GPU using SGLang. 
  }%
  \label{fig:teaser}%
  \vspace{-0.5em}
  \endgroup
}

\makeatletter
\renewcommand{\maketitle}{%
  \twocolumn[%
    \begin{adjustwidth}{0pt}{24pt}%
      \begin{flushleft}%
        {\raggedright \titlefont \@title\par}%
        \vskip11pt%
        {\raggedright \@author\par}%
        \vskip20pt%
      \end{flushleft}%
    \end{adjustwidth}%
    \abscontent
    \vskip16pt
    \teaserfigure
    \vskip20pt%
  ]%
  \thispagestyle{firststyle}%
}
\makeatother

\begin{document}
\maketitle

\begin{figure*}[t]
\centering
\includegraphics[width=0.99\linewidth]{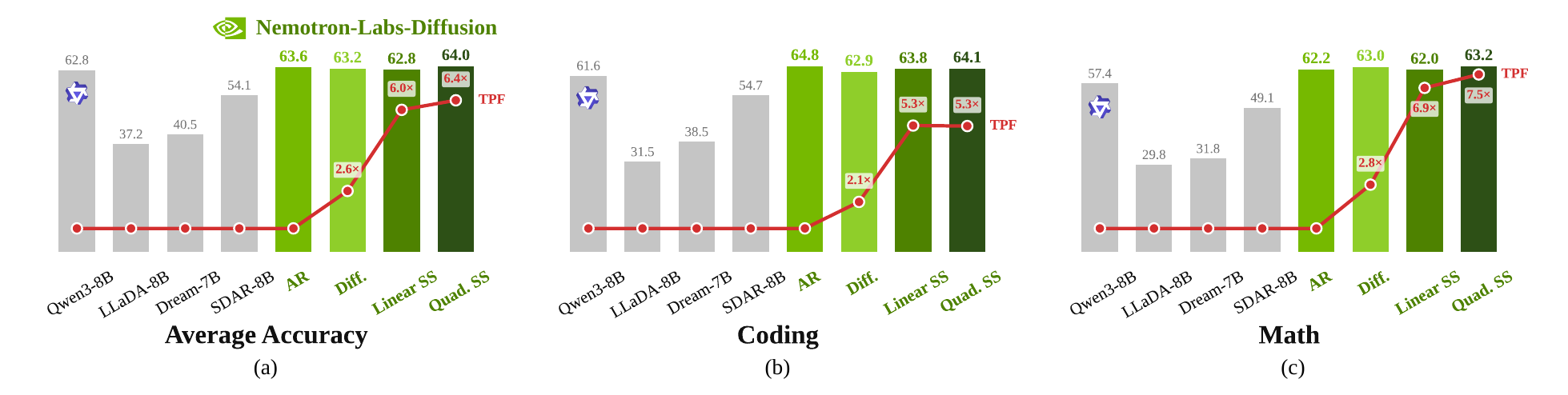} 
\vspace{-1em}
\caption{Benchmarking our \METHOD{}-8B (Instruct) against SOTA AR and diffusion instruct LMs across different benchmarks. (a) shows the average accuracy across all 10 tasks (HumanEval, MBPP, LiveCodeBench-CPP, GSM8K, Math500, AIME24, AIME25, GPQA, IFEval, MMLU) and the tokens per forward (TPF) of different models, while (b) and (c) show the average accuracy across coding (HumanEval, MBPP, LiveCodeBench-CPP) and math (GSM8K, Math500, AIME24, AIME25) domains, respectively.}
\label{fig:benchmark_bar}
\vspace{-1.5em}
\end{figure*}

\section{Introduction}
\label{sec:intro}

The strictly sequential, token-by-token decoding process of autoregressive (AR) language models (LMs) fundamentally limits their inference parallelism, resulting in resource under-utilization and low throughput, especially in low-batch-size deployment scenarios. Diffusion LMs~\cite{nie2025large,ye2025dream,cheng2025sdar} have recently emerged as a promising alternative, enabling parallel generation by decoding multiple tokens per forward pass. Nevertheless, diffusion LMs often lag behind AR models in accuracy and learning efficiency, requiring substantially more data to reach comparable performance~\cite{nie2024scaling}. 
A key reason is that diffusion training treats all token permutations uniformly~\cite{xue2025any}, rather than leveraging the strong left-to-right prior inherent in natural language. Moreover, existing diffusion LMs still lack clear advantages over multi-token prediction (MTP) methods and often fall behind them in practical efficiency–accuracy trade-offs.

These limitations raise three critical questions for understanding the role of diffusion LMs:

\textit{Q1: Should diffusion LMs compete with AR LMs, or can the two paradigms be harmonized?}

\textit{Q2: Can diffusion LMs provide a stronger acceleration mechanism than MTP methods?}

\textit{Q3: Does diffusion decoding have enough long-term potential to justify deeper exploration?}

Answering these questions is critical for judging the true promise of diffusion LMs and guiding their correct and wide adoption. This work studies these questions by \textbf{unifying AR and diffusion modeling within a single model} that preserves the strengths of AR LMs while exploring the benefits and potential of parallel decoding. The motivation is that AR and diffusion LMs might not be competing paradigms in which one should replace the other; instead, they can be mutually beneficial and unified within a single model by switching between causal and bidirectional attention. Specifically, AR models inherently learn to plan ahead for future tokens~\cite{samragh2025your}, and diffusion training can further enhance this capability. Conversely, preserving AR objectives during training injects strong left-to-right linguistic priors into diffusion modeling.

Based on this insight, we introduce \METHOD{}, a tri-mode LM that jointly optimizes diffusion and AR losses under a unified training framework. Our training scheme employs a global loss-averaging strategy that treats all tokens across batches equally to stabilize optimization. We further adopt a two-stage training procedure: we first strengthen AR capabilities to establish strong left-to-right linguistic priors, and then enable joint diffusion and AR training to fully integrate both objectives. The resulting models support tri-mode decoding as shown in Fig.~\ref{fig:teaser} (a): (1) AR decoding, (2) parallel diffusion-based decoding, which can be paired with a sampler optimized on sampling trajectories for improved parallelism, and (3) self-speculation, where diffusion drafts candidate tokens and AR predictions verify them.

We leverage this training scheme to deliver the \METHOD{} model family, including base, instruct, and vision-language variants at 3B/8B/14B scales. As shown in Fig.~\ref{fig:benchmark_bar}, our models outperform state-of-the-art (SOTA) open-source AR / diffusion LMs in both accuracy and inference speed across a wide range of benchmarks. For example, our \METHOD{}-8B delivers $6\times$ tokens per forward over Qwen3-8B while maintaining comparable or better accuracy on general benchmarks, translating to $4\times$ throughput on SPEED-Bench~\cite{abramovich2026speed} measured with SGLang on an NVIDIA GB200 GPU.

The results and analysis of our tri-mode LMs offer rich insights to answer the above questions. 
First, AR and diffusion LMs can be harmonized rather than treated as competing alternatives and unifying AR and diffusion objectives is a promising pathway: AR contributes strong next-token modeling and linguistic priors, while diffusion unlocks parallel generation without sacrificing benchmark performance. 
Beyond accuracy, the joint objective naturally enables self-speculation, allowing tri-mode LMs to adapt to different deployment regimes with different levels of concurrency: self-speculation is especially effective in low-concurrency settings, as shown in Fig.~\ref{fig:teaser} (b), while AR remains well suited for compute-bound high-concurrency scenarios. As such, tri-mode models can serve as drop-in replacements for conventional AR LMs, requiring no architectural or pipeline changes while offering consistently high throughput across deployment scenarios.

Our speed-of-light (SOL) analysis, which estimates the upper bound of diffusion decoding when equipped with an optimal sampler, shows that diffusion decoding has strong potential and substantial headroom beyond current parallel decoding methods. Specifically, diffusion-based decoding with an optimal sampler can correctly predict over $76.5\%$ more tokens per forward pass than the self-speculation mode, indicating that current samplers still leave a large fraction of the available parallelism unused. These results highlight the long-term promise of diffusion decoding.

Notably, we find that sampling tokens from the diffusion mode to approach the SOL upper bound remains an open challenge, and that the most effective approach is to verify the decoded tokens using the same model in AR mode. This observation motivates the aforementioned self-speculation mode, where diffusion generates high-quality multi-token drafts while AR verifies them within a single model, eliminating the need for the auxiliary prediction heads used by prior MTP methods~\cite{li2025eagle}. As shown in Fig.~\ref{fig:teaser} (c), this yields higher acceptance rates and more favorable trade-offs between system throughput and per-user throughput, making tri-mode LMs a stronger and more flexible alternative to existing MTP approaches.

We hope the above insights can shed light on the proper adoption of diffusion objectives in language modeling and on future directions that fully unlock the potential of diffusion decoding.

\noindent \textbf{Paper structure.} The rest of the paper is organized as follows. 
Sec.~\ref{sec:method_training} and Sec.~\ref{sec:tri-mode} detail our joint AR-diffusion training framework and tri-mode inference algorithms; Sec.~\ref{sec:sol} presents the speed-of-light analysis; Sec.~\ref{sec:model_family} and Sec.~\ref{sec:exp} introduce the \METHOD{} model family and present experiments, including comparisons between self-speculation and MTP; Sec.~\ref{sec:related-work} reviews related work and Sec.~\ref{sec:conclusion} concludes with insights and future directions.

\vspace{-0.5em}
\section{Tri-Mode LM Training}
\label{sec:method_training}

\subsection{Training Objectives}
\label{sec:objectives}

\textbf{Motivation.} We hypothesize that AR and diffusion objectives are complementary rather than competing. AR pretraining induces an implicit ability to plan ahead~\cite{samragh2025your}, which the diffusion objective further unlocks by forcing the model to reason about future tokens; in turn, the AR objective anchors diffusion training to the left-to-right structure of language and prevents wasted capacity on arbitrary token permutations. Therefore, we train \METHOD{} on a weighted combination of an AR next-token loss and a block-wise diffusion denoising loss.

\noindent \textbf{AR objective.}
For a token sequence $x$, the AR objective maximizes the likelihood under the left-to-right factorization:
\begin{equation}
\mathcal{L}_{\text{AR}}(\theta)
=
\mathbb{E}_{x \sim \mathcal{D}}
\left[
-\sum_{i=1}^{|x|}
\log p_{\theta}\!\left(x_i \mid x_{<i}\right)
\right].
\label{eq:lar}
\end{equation}

\begin{figure}[t]
\centering
\includegraphics[width=0.99\linewidth]{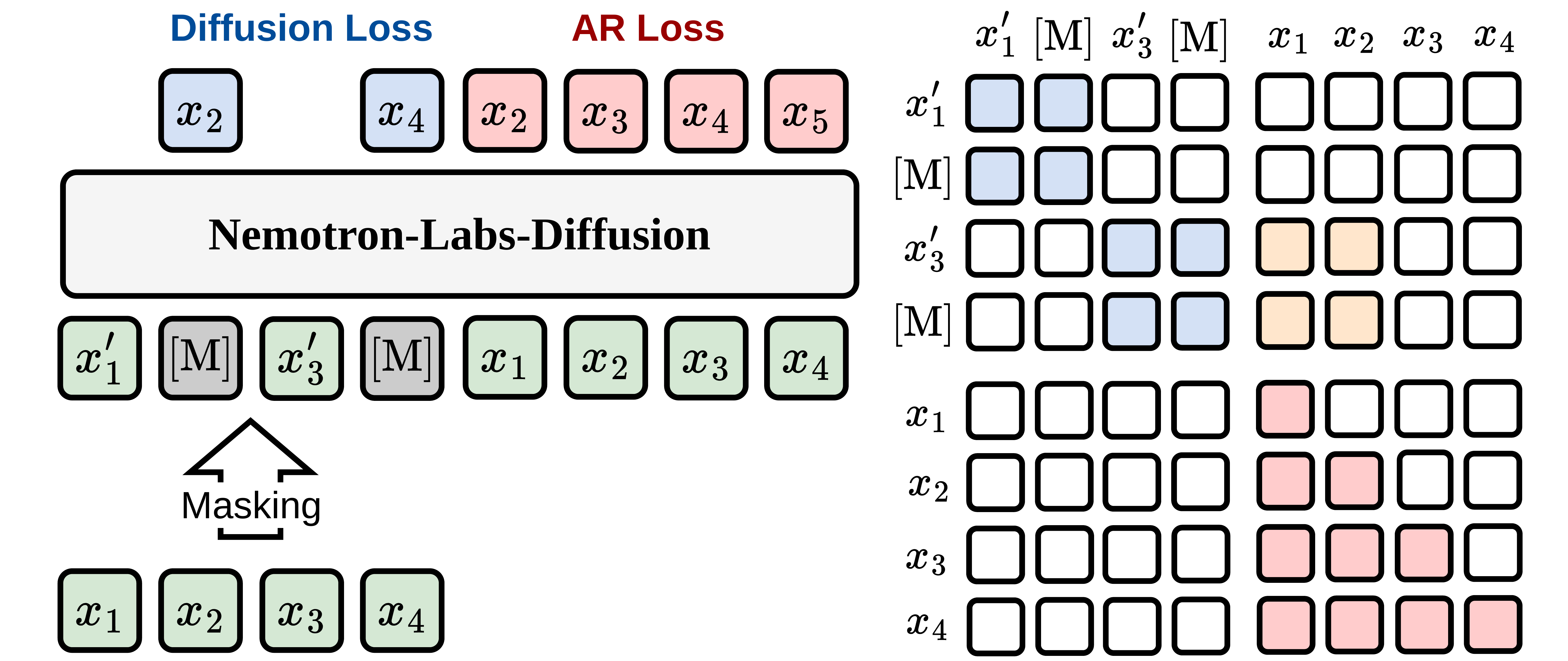} 
\vspace{-1em}
\caption{Visualizing the attention pattern of \METHOD{}, where \tinybox{maskblue} denotes attention among noisy tokens, \tinybox{maskorange} denotes attention from noisy tokens to clean-context tokens, and \tinybox{maskred} denotes attention within the clean context.
}
\label{fig:attention_pattern}
\vspace{-1em}
\end{figure}

\noindent \textbf{Diffusion objective.} As shown in Fig.~\ref{fig:attention_pattern},
we adopt the block-wise diffusion formulation~\cite{arriola2025block,fu2025efficient}, which partitions the sequence into $B$ contiguous blocks $\{x^b\}_{b=1}^{B}$ and trains the model to denoise one block at a time conditioned on its clean prefix. At noise level $t \sim \mathcal{U}[0,1]$, only the tokens in the current block are corrupted via a forward noising process $q$, i.e., $\tilde{x}_t^b \sim q(\cdot \mid x^b)$, while the prefix $x^{<b}$ remains clean:
\begin{equation}
\mathcal{L}_{\text{diff}}(\theta)
=
\mathbb{E}_{\substack{t \sim \mathcal{U}[0,1] \\
\tilde{x}_t^b \sim q(\cdot \mid x^b)}}
\left[
-\frac{1}{t}
\sum_{b=1}^{B}
\log p_{\theta}\!\left(x^b \mid \tilde{x}_t^b,\; x^{<b}\right)
\right].
\label{eq:diff_obj}
\end{equation}
This block-wise design is bidirectional within each block to enable parallel intra-block prediction, and causal across blocks so that previously generated blocks can reuse their KV cache during inference.

\noindent \textbf{Joint objective.}
We optimize a weighted combination of the two losses:
\begin{equation}
\mathcal{L}(\theta)
=
\mathcal{L}_{\text{AR}}(\theta)
+
\alpha \, \mathcal{L}_{\text{diff}}(\theta),
\label{eq:joint}
\end{equation}
where the AR loss has coefficient $1$ and $\alpha$ controls the strength of diffusion supervision. This design choice is motivated by the observation that the diffusion loss is often larger than the AR loss, and selecting an $\alpha$ that aligns the magnitudes of the two losses yields the best results, i.e., enabling diffusion-style parallel decoding while maintaining AR accuracy. We set $\alpha = 0.3$ across all training stages.

\begin{table*}[!t]
\centering
\caption{Ablation study of each training technique during continuous pretraining on 25B tokens.
}
\vspace{-0.5em}
\resizebox{\textwidth}{!}{
\begin{tabular}{cccccccc}
\toprule
\textbf{Technique} & \textbf{HumanEval} & \textbf{HumanEval+} & \textbf{MBPP} & \textbf{MBPP+} & \textbf{GSM8K} & \textbf{Minerva Math} & \textbf{Avg} \\ \midrule
Block-wise attention & 39.02 & 37.80 & 53.40 & 67.72 & 82.87 & 44.58 & 54.23 \\
+ Global Loss Avg & 42.07 & 39.02 & 56.20 & 71.69 & 83.78 & 45.36 & 56.35 \\
+ DP-rank Varying Masking Ratios & 45.12 & 43.29 & 55.80 & 70.90 & 81.58 & 45.66 & 57.06 \\
+ Two-stage training & 58.54 & 52.44 & 53.00 & 73.81 & 83.17 & 55.84 & 62.80 \\
+ AR loss & 64.02 & 57.93 & 65.60 & 80.95 & 86.73 & 66.44 & 70.28 \\ \bottomrule
\end{tabular}
}
\label{tab:ablation_base}
\vspace{-1em}
\end{table*}

\noindent \textbf{Two-stage training.}
To strengthen left-to-right priors and improve learning efficiency, we adopt a two-stage training strategy that first trains with the AR objective, which anchors the representation to the language's inherent left-to-right inductive bias, and then switches to the joint objective. In terms of Eq.~\ref{eq:joint}, Stage~1 sets $\alpha = 0$, reducing the optimization to the pure AR objective in Eq.~\ref{eq:lar}. In Stage~2, we turn on diffusion supervision by setting $\alpha$ to align the magnitudes of the two losses, as mentioned above, so that diffusion gradients complement rather than overwrite the AR priors.

\noindent\textbf{Global loss averaging.}
Since the diffusion objective involves randomly sampling masked tokens, different training examples may have different numbers of tokens contributing to the diffusion loss. As a result, the strategy for averaging token-wise losses matters, analogous to how the choice of aggregation in on-policy RL objectives (e.g., GRPO~\cite{shao2024deepseekmath} vs. DAPO~\cite{yu2025dapo}) can affect training stability. 
We consider two loss averaging strategies.
Let a batch contain $N$ sequences, each of length $L$, and let $\ell_{n,i}$ denote the token-level loss for token $i$ in sequence $n$ based on Eq.~\ref{eq:joint}.
One choice is to first average token losses within each sequence and then average them over sequences:
\begin{equation}
\mathcal{L}_{\text{seq-avg}}
=
\frac{1}{N}\sum_{n=1}^{N}
\Big(
\frac{1}{L}\sum_{i=1}^{L}\ell_{n,i}
\Big).
\label{eq:seqavg}
\end{equation}
Another choice is to treat all tokens across the batch equally and globally average over the $NL$ token losses:
\begin{equation}
\mathcal{L}_{\text{global}}
=
\frac{1}{NL}\sum_{n=1}^{N}\sum_{i=1}^{L}\ell_{n,i}.
\label{eq:globalavg}
\end{equation}

While Eq.~\ref{eq:seqavg} and Eq.~\ref{eq:globalavg} coincide when every sequence has the same number of loss-contributing tokens, they differ once masking yields variable numbers of noisy tokens across samples, which is common in the diffusion objective. In particular, in Eq.~\ref{eq:diff_obj}, the loss includes a $\tfrac{1}{t}$ reweighting, and the number of noisy tokens is approximately proportional to $t$. When $t$ is small, each noisy token tends to carry a larger weight (due to $\tfrac{1}{t}$), but there are fewer such tokens in the sample. 
Sequence-wise averaging can therefore amplify the influence of these small-$t$ samples: their per-token losses are larger, yet the per-sequence normalization assigns them the same weight as other samples, increasing batch-to-batch fluctuations and gradient variance. 
In contrast, global averaging effectively weights each training example in proportion to its number of contributing tokens, preventing samples with only a few highly weighted noisy tokens from disproportionately influencing the batch loss.

\subsection{Attention Pattern}
\label{sec:attn}

Following~\cite{arriola2025block}, at training time we use a dual-stream input by concatenating a corrupted/noised view and a clean view of the same sequence, and apply a structured attention pattern, as shown in Fig.~\ref{fig:attention_pattern}. 

The \textit{Noisy$\rightarrow$Noisy} and \textit{Noisy$\rightarrow$Clean} parts follow the standard block diffusion design~\cite{arriola2025block}: we partition the sequence into $B$ contiguous blocks $\{x^b\}_{b=1}^{B}$; in the noisy stream, tokens attend bidirectionally within each block and causally across blocks; and for denoising block $b$, noisy tokens additionally attend to the clean-prefix blocks $x^{<b}$ in the clean stream to achieve clean-context conditioning.

The key difference lies in the \textit{Clean$\rightarrow$Clean} mask. Prior designs~\cite{arriola2025block, fu2025efficient, wu2025fast2} allow the clean stream to attend to future tokens using block-wise attention. In contrast, we enforce a strictly causal mask within the clean stream~\cite{gat2025set,samragh2025your}. This enables us to compute the AR objective on $x$ in this clean-context part together with the diffusion objective on $\tilde{x}_t$ in the same forward-backward pass, without label leakage.

\noindent \textbf{Relationship with prior works.} Our attention pattern follows the pioneering work of \cite{gat2025set}, which also performs joint diffusion and AR training. The key differences in our work lie in (1) proposing tri-mode inference, particularly self-speculation decoding, along with post-training enhancements for improved parallelism, including the samplers and LoRA-enhanced drafters introduced in Sec.~\ref{sec:tri-mode}; (2) the overall training pipeline used to develop the full model family described in Sec.~\ref{sec:model_family}; and (3) the systematic studies and SOL analysis conducted to address critical questions regarding the true potential of diffusion LMs.

\begin{figure*}[t!]
\centering
\vspace{-0.5em}
\includegraphics[width=0.92\linewidth]{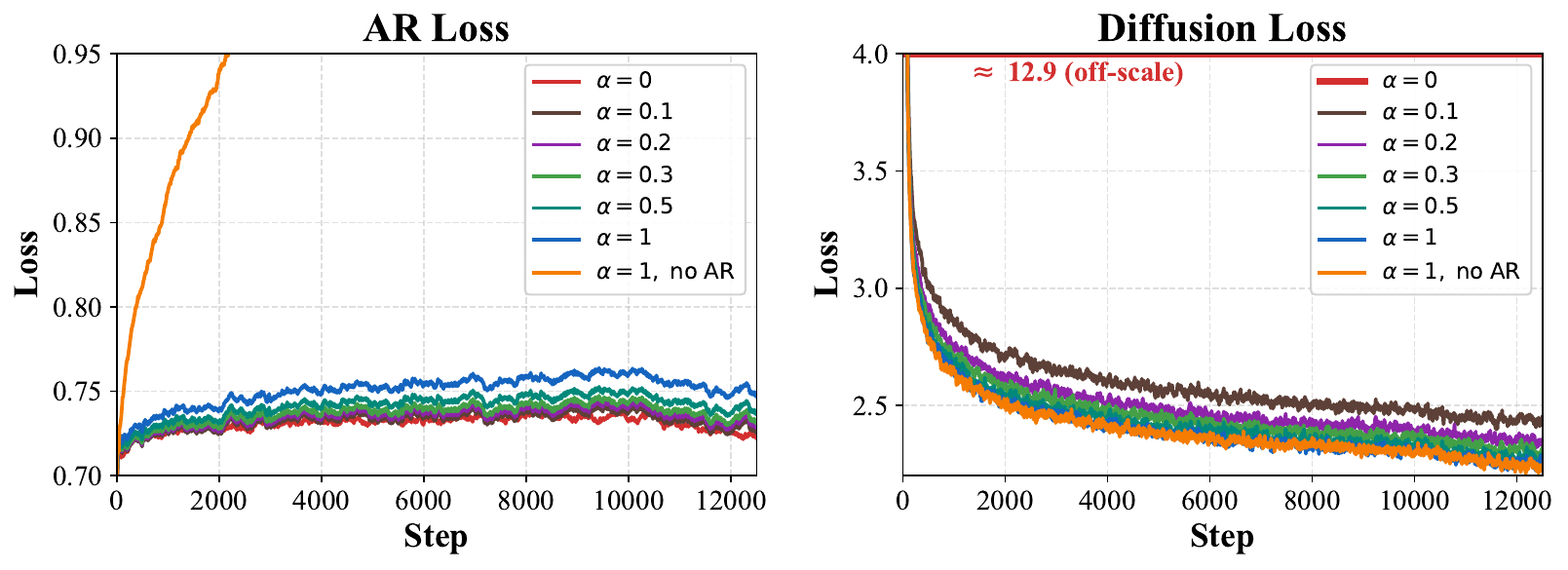} 
\vspace{-1em}
\caption{
Visualizing the evolution of the AR and diffusion losses across training steps under different diffusion loss coefficients $\alpha$. The AR loss coefficient is set to 1 by default across all settings, except in the `no AR' setting, where the AR loss is removed and only the diffusion loss is used.
}
\label{fig:loss_weight_curve}
\vspace{-1em}
\end{figure*}

\vspace{-1em}
\subsection{Ablation Study on Training Techniques}
\label{sec:exp_training_technique}

We ablate the contribution of each training technique by progressively adding them during continuous pretraining on 25B tokens, starting from the official Ministral3-8B base model. Detailed training/evaluation settings will be elaborated in Sec.~\ref{sec:base_model_setting} and Sec.~\ref{sec:exp_base}.
All models are evaluated in diffusion mode on coding and math benchmarks.

\noindent \textbf{Observations.}
As shown in Tab.~\ref{tab:ablation_base}, we progressively add each technique and observe that
(1) \textit{block-wise attention} serves as the baseline at 54.23\% average accuracy, following the setting of~\cite{fu2025efficient, cheng2025sdar} and providing a functional diffusion LM;
(2) \textit{global loss averaging} improves the average by 2.12\%, confirming that treating all tokens equally across the batch reduces gradient variance from variable masking ratios, as analyzed in Sec.~\ref{sec:objectives};
(3) \textit{DP-rank varying masking ratios}, which applies different noise levels across data-parallel ranks, further improves the average by 0.71\%;
(4) \textit{two-stage training}, which provides a better AR initialization with 1T-token AR objective training, yields a substantial 5.74\% gain, implying that stronger AR initialization can enable better future planning and ease AR-to-diffusion conversion;
(5) \textit{the addition of AR loss} contributes the largest single improvement of 7.48\%, significantly boosting diffusion decoding abilities, echoing our analysis in Sec.~\ref{sec:method_training}.

Cumulatively, the full pipeline improves the baseline by 16.05\% in average accuracy, with the AR loss and two-stage training contributing the most. This validates our core insight that preserving the AR objective during diffusion training anchors the model to linguistically coherent trajectories and is a critical factor for achieving strong diffusion LM accuracy.

\begin{table}[t]
\centering
\caption{Impact of diffusion loss weight $\alpha$ during 25B-token continuous pretraining.}
\vspace{-0.5em}
\resizebox{\columnwidth}{!}{
\begin{tabular}{ccccccccc}
\toprule
\textbf{$\alpha$} & \textbf{Mode} & \shortstack{\textbf{Human}\\\textbf{Eval}} & \shortstack{\textbf{Human}\\\textbf{Eval+}} & \textbf{MBPP} & \textbf{MBPP+} & \textbf{GSM8K} & \shortstack{\textbf{Minerva}\\\textbf{Math}} & \textbf{Avg} \\
\midrule
\multirow{2}{*}{0.1} 
& Diff. & 56.71 & 51.83 & 64.80 & 81.22 & 87.64 & 67.02 & 68.20 \\
& AR    & 58.54 & 53.05 & 64.60 & 80.42 & 87.64 & \textbf{68.02} & 68.71 \\ 
\midrule
\multirow{2}{*}{0.2} 
& Diff. & 60.37 & 54.27 & 63.40 & 80.69 & 86.43 & 66.58 & 68.62 \\
& AR    & 60.98 & 57.93 & \textbf{66.40} & \textbf{83.60} & 87.04 & 67.04 & 70.50 \\ 
\midrule
\multirow{2}{*}{\textbf{0.3}} 
& Diff. & 61.59 & \textbf{58.54} & 64.60 & 80.42 & 87.64 & 65.86 & 69.77 \\
& AR    & \textbf{62.80} & 57.93 & 65.80 & 82.54 & \textbf{87.79} & 66.84 & \textbf{70.62} \\ 
\midrule
\multirow{2}{*}{0.5} 
& Diff. & 59.76 & 54.27 & 64.40 & 80.16 & 87.11 & 66.98 & 68.78 \\
& AR    & 58.54 & 53.05 & 65.40 & 82.80 & 86.81 & 67.14 & 68.96 \\ 
\midrule
\multirow{2}{*}{1.0} 
& Diff. & 56.10 & 48.78 & 65.00 & 80.69 & 84.91 & 66.12 & 66.93 \\
& AR    & 54.27 & 50.61 & 64.80 & 80.69 & 86.58 & 66.36 & 67.22 \\ 
\bottomrule
\end{tabular}
}
\label{tab:ablation_alpha}
\vspace{-1.5em}
\end{table}

\begin{table*}[!t]
\centering
\vspace{-0.5em}
\caption{AR-mode accuracy with and without the diffusion loss ($\alpha{=}0.3$). \textit{Base}: 25B-token continuous pretraining from Ministral3-8B. \textit{Instruct}: further SFT on 45B tokens.}
\vspace{-0.5em}
\resizebox{\textwidth}{!}{
\begin{tabular}{cccccccccccc}
\toprule
\multicolumn{12}{c}{\textit{Base Model}} \\ \midrule
\textbf{Training} & \textbf{HumanEval} & \textbf{HumanEval+} & \textbf{MBPP} & \textbf{MBPP+} & \textbf{GSM8K} & \textbf{Minerva Math} & \textbf{MMLU} & \textbf{Hellaswag} & \textbf{PIQA} & \textbf{Winogrande} & \textbf{Avg} \\ \midrule
AR only & 60.37 & 56.10 & 67.00 & 81.48 & 87.64 & 67.90 & 76.34 & 76.54 & 79.71 & 71.98 & 72.50 \\
+ Diff. loss & 62.80 & 57.93 & 65.80 & 82.54 & 87.79 & 66.84 & 75.99 & 76.59 & 79.82 & 70.32 & 72.64 \\ \midrule
\multicolumn{12}{c}{\textit{Instruct Model}} \\ \midrule
\textbf{Training} & \textbf{GPQA} & \textbf{IFEval} & \textbf{HumanEval} & \textbf{MBPP} & \textbf{Math500} & \textbf{GSM8K} & \textbf{AIME24} & \textbf{AIME25} & \textbf{MMLU} & \textbf{LCB-CPP} & \textbf{Avg} \\ \midrule
AR only & 44.44 & 71.66 & 82.93 & 83.60 & 87.80 & 93.63 & 36.67 & 26.67 & 79.77 & 24.61 & 63.18 \\
+ Diff. loss & 44.44 & 68.65 & 80.49 & 85.19 & 88.00 & 94.01 & 33.33 & 33.33 & 79.85 & 28.85 & 63.61 \\ \bottomrule
\end{tabular}
}
\label{tab:ablation_ar_impact}
\vspace{-1em}
\end{table*}

\subsection{Mutual Impact of AR/Diffusion Losses}
\label{sec:exp_mutual}

In this subsection, we examine whether the AR and diffusion objectives compete for model capacity or reinforce each other: the impact of adding the AR loss on the diffusion mode, and the impact of adding the diffusion loss on AR-mode accuracy.

\noindent \textbf{AR loss boosts diffusion accuracy.}
As studied in Sec.~\ref{sec:exp_training_technique} and Tab.~\ref{tab:ablation_base}, AR loss can significantly boost diffusion accuracy. As a complement to this study, we also vary $\alpha$ in Eq.~\ref{eq:joint} during 25B-token continuous pretraining on top of the two-stage training setting in Tab.~\ref{tab:ablation_alpha}. We observe that both modes peak at $\alpha{=}0.3$. This implies that the two modes do not necessarily compete with each other or achieve the best performance at the two extremes; instead, there exists a sweet spot where both are well harmonized. Similarly, no value of $\alpha$ in $[0.1, 0.5]$ improves one mode at the expense of the other, and the two objectives rise and fall together, indicating that they are complementary rather than competing for model capacity.

We also visualize the training loss curves in Fig.~\ref{fig:loss_weight_curve}. We observe that the aforementioned setting $\alpha{=}0.3$, which achieves the best accuracy, provides a good balance between the two losses. Setting $\alpha$ too small or too large leads to increased diffusion or AR loss, respectively. In addition, without the AR or diffusion loss, the corresponding AR or diffusion capabilities are degraded or lost.

\noindent \textbf{Diffusion loss preserves AR accuracy.}
To study whether diffusion loss can hurt or preserve AR accuracy, we compare models trained w/o and w/ the diffusion loss ($\alpha{=}0.3$) under two settings: continuous pretraining on 25B tokens on top of the two-stage training setting in Tab.~\ref{tab:ablation_base}, and further SFT on 45B tokens, following training/evaluation settings in Sec.~\ref{sec:instruct_model_setting} and  Sec.~\ref{sec:exp_instruct}. We ensure that all settings are trained on the same number of tokens.

As shown in Tab.~\ref{tab:ablation_ar_impact}, we observe that:
(1) In both settings, the average AR accuracy is preserved or slightly boosted, with 0.14\% and 0.43\% improvements for the base and instruct models, respectively, indicating that diffusion training, when properly integrated, can enhance the future prediction abilities of the AR mode, similar to observations in DeepSeek-V3~\cite{deepseekai2024deepseekv3technicalreport};
(2) At the per-benchmark level, the instruct model shows gains on coding and math benchmarks, e.g., 4.24\% higher on LCB-CPP and 1.59\% higher on MBPP, but drops on IFEval (3.01\% lower) and HumanEval (2.44\% lower), suggesting that strict instruction-following compliance is slightly affected by the diffusion objective.

These results, together with the $\alpha$ sensitivity analysis above, support the conclusion that joint AR–diffusion training is not a zero-sum trade-off: the diffusion loss enables parallel decoding modes (Sec.~\ref{sec:tri-mode}) at negligible cost to AR-mode accuracy, and the two objectives share a common optimal operating point.

\begin{figure*}[t]
\centering
\includegraphics[width=0.99\linewidth]{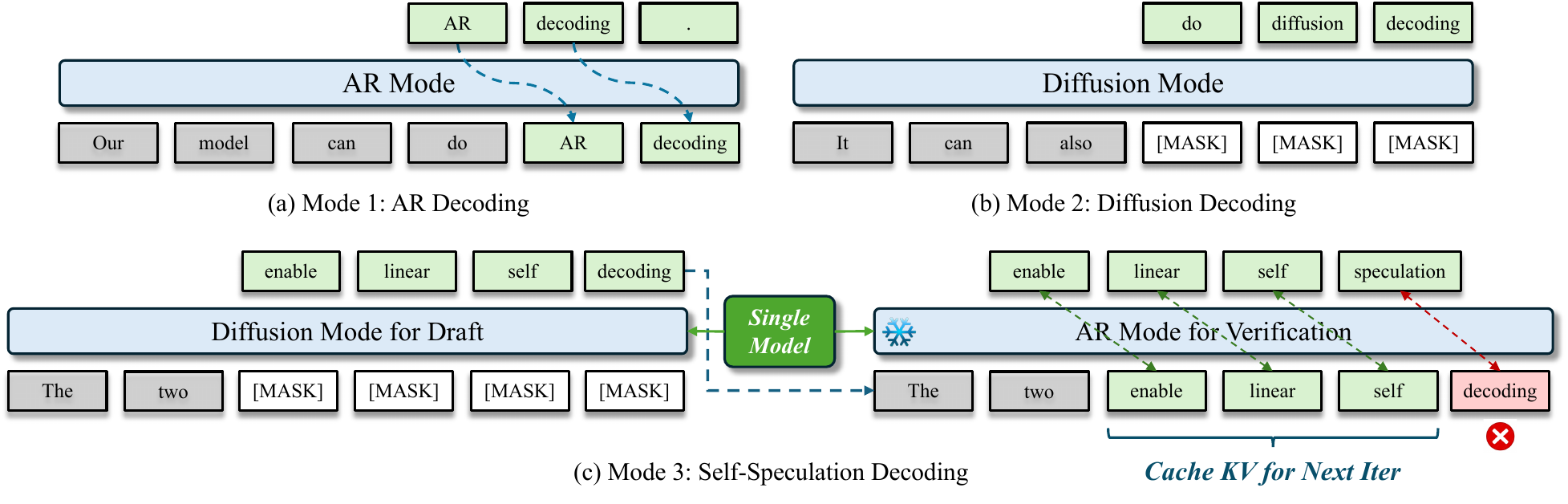} 
\vspace{-0.5em}
\caption{
Visualizing tri-mode inference: (a) left-to-right AR decoding, (b) parallel diffusion decoding, and (c) linear self-speculation decoding (quadratic self-speculation is visualized in Fig.~\ref{fig:quadratic_speculation}).
}
\label{fig:tri_mode_inference}
\vspace{-1em}
\end{figure*}

\section{Tri-Mode LM Inference}
\label{sec:tri-mode}

The joint AR and diffusion training enables decoding in three modes: AR, diffusion, and self-speculation decoding, as shown in Fig.~\ref{fig:tri_mode_inference}.

\subsection{Mode 1: AR Decoding}
\label{sec:ar_mode_inference}

Tri-mode LMs fully preserve standard left-to-right generation: at step $i$, they sample $x_i \sim p_{\theta}(\cdot \mid x_{<i})$ with causal attention. This mode is preferred when serving with high concurrency.

\subsection{Mode 2: Block-wise Diffusion Denoising}
\label{sec:diff_mode_inference}

\noindent \textbf{Confidence-based sampling.}
Following~\cite{wu2025fast2,fu2025efficient}, the diffusion decoding mode proceeds block by block. For the current block, we initialize its positions as mask tokens and iteratively denoise multiple tokens in parallel per step based on a confidence threshold~\cite{wu2025fast}. When a block is completed, its KV cache will be refreshed, and decoding proceeds to the next block.

\noindent \textbf{Sampling with a trained sampler.}
A fixed confidence threshold is an implicit signal that is not explicitly optimized during training. We therefore train a lightweight sampler that, for every masked position in the current block, predicts whether the top-$1$ prediction at the current denoising step is \emph{correct}. Here, \emph{correct} means that the decoded token matches the token that will eventually be committed at this position when decoding only the highest-confidence token at each step.
At inference, we commit positions whose predicted probability from the sampler exceeds a predefined threshold, which trades off TPF against per-token error rate. The sampler can be viewed as a learned classifier to approach the greedy-acceptance criterion that we later analyze in Sec.~\ref{sec:sol}.
The sampler architecture, feature engineering, and training trajectory data collection are detailed in Appendix~\ref{sec:appendix_sampler}.

\vspace{-0.5em}
\subsection{Mode 3: Self-Speculation Decoding}
\label{sec:ss_mode_inference}

\noindent \textbf{Linear self-speculation.}
The simplest self-speculative mode separates diffusion-based drafting and AR-based verification into two forward passes. Let $[x_1,\ldots,x_n]$ denote the currently verified prefix and let $k$ be the speculative width.

\noindent \underline{Drafting with diffusion.}
We append $k$ mask tokens to the verified prefix, forming the input $[x_1,\ldots,x_n, m_1,\ldots,m_k]$. The model denoises all $k$ mask positions in parallel using the diffusion pathway, producing draft tokens $\{\hat{x}_{n+1},\ldots,\hat{x}_{n+k}\}$.

\noindent \underline{Verification with AR.}
We then run a second forward pass over the draft tokens $[\hat{x}_{n+1},\ldots,\hat{x}_{n+k}]$ with causal attention, again reusing the prefix KV cache. The AR logits at each position yield next-token predictions $\{x^{\text{AR}}_{n+j}\}_{j=1}^{k}$. We accept the longest prefix of draft tokens that passes the verification criterion (e.g., $x^{\text{AR}}_{n+j} = \hat{x}_{n+j}$) and commit the accepted tokens to the verified prefix. As in standard speculative decoding~\cite{leviathan2023fast}, the AR prediction at the first rejected position provides one additional verified token, so each step produces between 1 and $k{+}1$ tokens. Note that both the drafting and verification passes can reuse the cached prefix KVs from prior verified steps.

\noindent \textbf{Enhance linear self-speculation w/ LoRA.}
We further enhance linear self-speculation by tuning a LoRA adapter~\cite{hu2022lora} on top of the diffusion draft pathway to better align its drafts with the AR verifier, thereby extending the accepted prefix length per step. We apply LoRA only to the $o_{\text{proj}}$ layer of the attention module (rank $128$, $\alpha{=}512$, $\sim$$36$M trainable parameters/$\sim$$0.4\%$ of the backbone), leaving the AR pathway unchanged. The training loss combines an LK-hybrid distribution-matching term~\cite{samarin2026lk} with a token-level cross-entropy term, both applied to the accepted prefix plus the first rejected position of each draft block, as shown in Fig.~\ref{fig:lora_loss} in Appendix~\ref{sec:appendix_linear_ss}.

\noindent \underline{Drafter--verifier setup.}
For each position $j \in \{1, \ldots, k\}$ in the draft block, the LoRA-augmented drafter produces logits $z^d_j \in \mathbb{R}^{|\mathcal{V}|}$ over the vocabulary $\mathcal{V}$, while the frozen AR verifier produces target logits $z^t_j$ on the same context. We define the temperature-scaled distributions
\[
q_j \;=\; \mathrm{softmax}(z^d_j / \tau), \qquad p_j \;=\; \mathrm{softmax}(z^t_j / \tau),
\]
with $\tau{=}3.0$. The target $p_j$ is treated as fixed (stop-gradient), so only the drafter $q_j$ carries gradient through the LoRA parameters.

\noindent \underline{Active position mask: ``accepted + 1''.}
As shown in Fig.~\ref{fig:lora_loss}, both loss terms are computed only on the \emph{accepted prefix plus the first rejected position}. Letting $j^{*}$ denote the position of the first mismatch (the smallest $j$ with $\hat{x}_{n+j} \neq x^{\mathrm{AR}}_{n+j}$), the active set is $\mathcal{A} = \{1, \ldots, j^{*}\}$ when there is a rejection in the block, and $\mathcal{A} = \{1, \ldots, k\}$ otherwise; all other positions are masked out and contribute neither to the loss numerator nor to its denominator. This mask is essential because, at inference, the verifier's KV cache is rebuilt at the rejection point: logits at positions $j > j^{*}$ are conditioned on a continuation the deployed loop never observes, so training on them would bias the drafter toward a counterfactual distribution.

\noindent \underline{LK-hybrid distribution-matching loss.}
The distribution-matching term adapts the LK-hybrid loss of~\cite{samarin2026lk} to a truncated top-$K$ support. We retain the union of the top-$K$ token indices, $\mathcal{U}_j = \mathcal{S}^t_j \cup \mathcal{S}^d_j$, where $\mathcal{S}^t_j$ (resp.\ $\mathcal{S}^d_j$) is the set of $K$ indices with the largest probability under $p_j$ (resp.\ $q_j$). Setting $K{=}200$ gives $|\mathcal{U}_j| \le 2K = 400$. We zero both distributions outside $\mathcal{U}_j$ and renormalize to obtain $\tilde p_j$ and $\tilde q_j$. Truncation to the union avoids the $\mathrm{KL}(\tilde p_j \,\|\, \tilde q_j) = \infty$ catastrophe of full-vocabulary KL. The per-position hybrid loss is
\begin{equation}
\mathcal{L}^{\mathrm{LK}}_j \;=\; \lambda_j \cdot \mathrm{KL}(\tilde p_j \,\|\, \tilde q_j) \;+\; (1-\lambda_j) \cdot \tfrac{1}{2} \sum_{v \in \mathcal{U}_j} \bigl|\tilde p_j(v) - \tilde q_j(v)\bigr|,
\label{eq:lk_hybrid}
\end{equation}
where the right-hand total-variation (TV) term equals $1 - \alpha_j$, with $\alpha_j = \sum_{v \in \mathcal{U}_j} \min\bigl(\tilde p_j(v), \tilde q_j(v)\bigr)$ the standard speculative-decoding acceptance probability~\cite{leviathan2023fast}---the probability that a token sampled from $\tilde q_j$ is accepted by the speculative-decoding rejection rule against $\tilde p_j$. The adaptive coefficient $\lambda_j = \exp(-\eta \cdot \mathrm{sg}[\alpha_j])$ with $\eta{=}0.5$ makes the loss behave like the (forward) KL early in training (when $\alpha_j \approx 0$ and $\lambda_j \approx 1$, providing a stronger distribution-matching gradient) and like TV as the drafter approaches the verifier ($\alpha_j \to 1$ and $\lambda_j \to e^{-\eta} \approx 0.6$, directly minimizing the acceptance-rate gap).

\begin{figure*}[!t]
\centering
\vspace{-0.5em}
\includegraphics[width=0.95\linewidth]{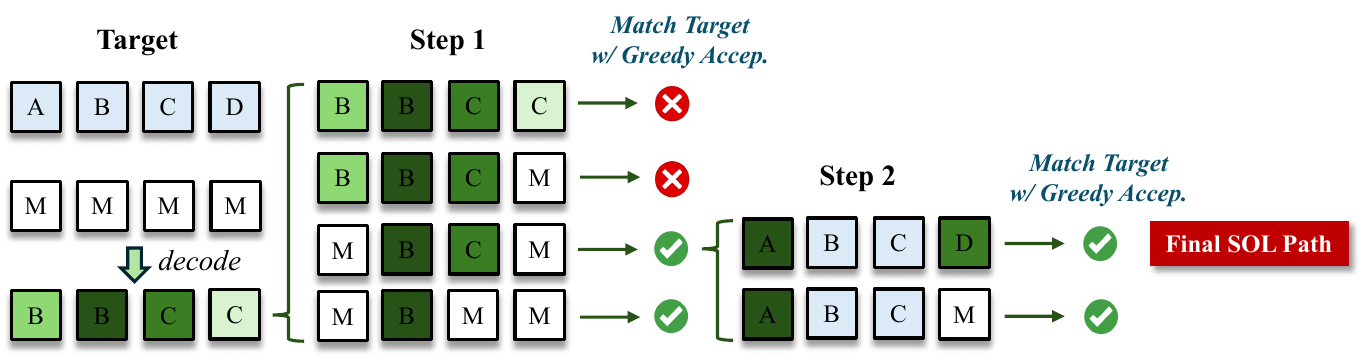} 
\vspace{-0.5em}
\caption{
An example of using the recursive dynamic compaction method to identify the SOL path.}
\label{fig:sol_method}
\vspace{-1em}
\end{figure*}

\noindent \underline{Cross-entropy term.}
The LK-hybrid term matches the full top-$K$ output distribution. We additionally apply a token-level cross-entropy at each active position against the verifier's argmax target $y_j = \arg\max_v z^t_j(v)$ on the same union support:
\begin{equation}
\ell_j \;=\;
\begin{cases}
-\log\, q^{(\mathcal{U}_j)}_j(y_j) & \text{if } y_j \in \mathcal{U}_j, \\
0 & \text{otherwise},
\end{cases}
\label{eq:lora_ce}
\end{equation}
where $q^{(\mathcal{U}_j)}_j$ is the drafter softmax restricted to $\mathcal{U}_j$ with $\tau{=}1.0$. The cross-entropy provides a strong teacher-forcing signal toward the verifier's modal token, complementing the soft distribution-matching of LK. When the truncation excludes $y_j$ (rare, $<2\%$ at $K{=}200$), $\ell_j$ is set to zero; that position is also dropped from the CE denominator below (Eq.~\ref{eq:lora_means}), so occasional truncation misses cannot blow up the loss.

\noindent \underline{Total loss and training-time drafter sampling.}
The two terms are aggregated as masked means over the active positions of each draft block:
\begin{equation}
\mathcal{L}_{\mathrm{LK}} \;=\; \frac{1}{|\mathcal{A}|} \sum_{j \in \mathcal{A}} \mathcal{L}^{\mathrm{LK}}_j,
\qquad
\mathcal{L}_{\mathrm{CE}} \;=\; \frac{\sum_{j \in \mathcal{A}} \ell_j}{\sum_{j \in \mathcal{A}} \mathbb{1}\{y_j \in \mathcal{U}_j\}}.
\label{eq:lora_means}
\end{equation}
These per-block means are further averaged across the inner training batch. The total loss is
\begin{equation}
\mathcal{L} \;=\; \lambda_{\mathrm{KL}} \cdot \mathcal{L}_{\mathrm{LK}} \;+\; \lambda_{\mathrm{CE}} \cdot \mathcal{L}_{\mathrm{CE}},
\qquad \lambda_{\mathrm{KL}} = \lambda_{\mathrm{CE}} = 1.
\label{eq:lora_total}
\end{equation}
At training time, $90\%$ of the inner-batch slots draw their drafter tokens by sampling from $\mathrm{softmax}(z^d_j / T_{\mathrm{draft}})$ with $T_{\mathrm{draft}}{=}1.0$; the rest stay greedy. Sampling exposes the LK gradient to a broader empirical distribution of drafter outputs and yields adapters that remain robust when the verifier is itself sampled at inference.

\subsection{Variant: Quadratic Self-Speculation}
\label{sec:quad_ss}

A variant of linear self-speculation is quadratic self-speculation, which leverages quadratic decoding~\cite{samragh2025your} with single-forward drafting and verification, following the same process as~\cite{liu2025tidar}.
This decoding scheme prepares for the worst case by predicting the next block at all possible acceptance positions with a quadratic cost. Specifically, quadratic self-speculation performs speculative drafting and verification simultaneously within a single forward pass by using a structured attention mask, where causal predictions verify previous drafts while parallel diffusion predictions generate new draft tokens for the next iteration. The interleaved quadratic layout ensures that each iteration consistently produces $k$ speculative tokens even when verification terminates early due to mismatches. In addition to standard AR-based verification, our tri-mode model further supports an AR-diffusion ensemble verifier that combines causal and diffusion predictions through weighted interpolation. More details are provided in Appendix~\ref{sec:appendix_quad_ss} and Fig.~\ref{fig:quadratic_speculation}.

\section{Speed-of-Light Analysis}
\label{sec:sol}
\vspace{-0.5em}

We conduct a speed-of-light (SOL) analysis to quantify the maximum acceptance rate / token-per-forward achievable by the diffusion mode. We apply this analysis to the diffusion mode of \METHOD{}-8B delivered in Sec.~\ref{sec:diff_mode_inference}. The SOL ceiling tells us how much intrinsic parallelism the current confidence-based sampling is leaving on the table. SOL is computed entirely within the diffusion model, i.e., no AR verifier is involved, so it isolates the diffusion model's own parallel-decoding capability and provides a reference ceiling for any scheme that targets its converged output. Compared with linear self-speculation in Sec.~\ref{sec:tri-mode}, which only commits a contiguous prefix of the draft and is therefore truncated at the first rejection, diffusion-mode decoding can commit \emph{any} subset of masked positions per pass.

\begin{figure*}[t]
\centering
\vspace{-0.5em}
\includegraphics[width=\linewidth]{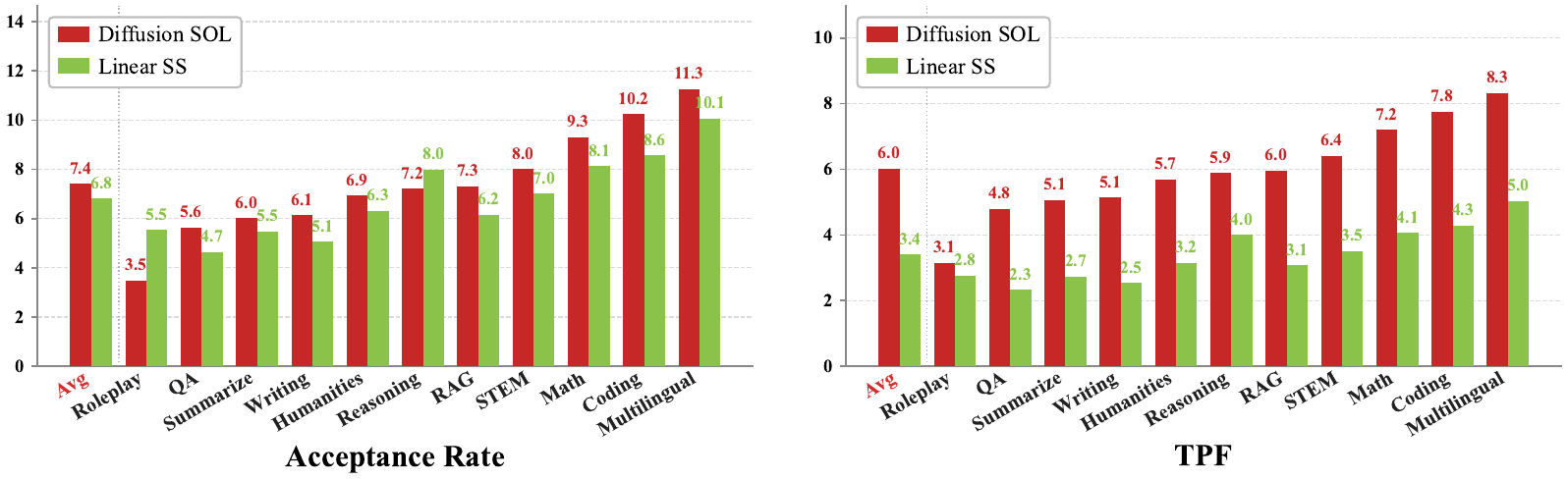} 
\vspace{-2em}
\caption{
Visualizing the acceptance rate and TPF across different SPEED-Bench categories for diffusion SOL and linear self-speculation. The average metrics across all categories are highlighted in red.
}
\label{fig:radar_sol_ss}
\vspace{-1em}
\end{figure*}

\vspace{-0.5em}
\subsection{Diffusion SOL Construction}

\noindent \textbf{Oracle target via serial denoising.}
We first define the diffusion model's converged output for each block of length $B$. Let $f_\theta$ denote the diffusion model, which on a partially masked input outputs a categorical distribution over the vocabulary at every masked position; let $\texttt{[M]}$ denote the mask token. Starting from an all-mask input $\mathbf{x}^{(0)} = \texttt{[M]}^B$, at each step we identify the masked position whose output distribution has the highest peak probability (across positions and vocabulary), commit its argmax to that position, and re-evaluate $f_\theta$ on the resulting partially-unmasked sequence; we repeat until all $B$ positions are filled. This \emph{serial denoising} procedure uses exactly $B$ forward passes—one position committed per pass—and yields a target sequence $\mathbf{t} = (t_1, \ldots, t_B)$ that the model would converge to in the absence of any parallel commits. The SOL acceptance ratio is then the average TPF needed to reproduce $\mathbf{t}$ from the same all-mask input under a parallel scheme.

\noindent \textbf{Greedy parallel acceptance.}
The simplest parallel scheme is greedy acceptance. At iteration $k$, the model produces argmax predictions $\hat{\mathbf{t}}^{(k)}$ for every position from the current input $\mathbf{x}^{(k)}$, and we commit every masked position whose prediction matches the serial target, $\mathcal{A}^{(k)} = \{j : x^{(k)}_j = \texttt{[M]} \;\land\; \hat{t}^{(k)}_j = t_j\}$; if no position matches, we commit the single highest-confidence position as a fallback. After $K$ iterations the block is fully unmasked and the realized TPF is $B/K$. Greedy is fast but is not always exact: each committed token becomes part of the context for the next forward pass, and committing several context-dependent tokens at once can shift the conditional distribution at the remaining positions away from $\mathbf{t}$. This motivates a second scheme that recovers $\mathbf{t}$ exactly on every block.

\noindent \textbf{Recursive dynamic compaction.}
To recover $\mathbf{t}$ exactly on every block, we replace greedy acceptance with a strategic search for the largest \emph{safe} subset of matching positions, as shown in Fig.~\ref{fig:sol_method}. At each iteration, we rank the $N$ matched positions by model confidence as $(p_1, \ldots, p_N)$ and search for the largest prefix $\{p_1, \ldots, p_k\}$ whose commit is safe—where ``safe'' means that continuing decoding on the remaining positions still arrives at $\mathbf{t}$. Each safety check itself runs the decoder one level shallower (greedy acceptance) under a simulation budget of $5000$ forward passes per block, which is what makes the scheme \emph{recursive}; if the budget is exceeded, the candidate is treated as unsafe and the binary search shrinks the prefix. Because the top-$1$ match is always safe (committing one position is no different from a serial step), the scheme commits at least one position per iteration and its TPF dominates greedy whenever greedy is exact, while recovering exact-match cases that greedy misses. We use this scheme to report SOL throughout this section; greedy serves as a fast lower-bound proxy.

\vspace{-0.5em}
\subsection{SOL Evaluation on SPEED-Bench}
\vspace{-0.5em}

We apply recursive dynamic compaction to $713$ SPEED-Bench~\cite{abramovich2026speed} samples spanning 11 categories on the diffusion mode of \METHOD{}-8B (instruct), sweeping the block length $B \in \{4, 8, 16, 32\}$. 
We also report the average accuracy achieved by SOL under different block lengths on the 10 instruct LM benchmarks in Sec.~\ref{sec:exp_instruct} to understand their impact.

\noindent \textbf{Observations on diffusion SOL.}
From Tab.~\ref{tab:recd1-speedbench}, we observe that
(1) The diffusion mode exhibits substantial intrinsic parallelism: the SOL acceptance rate reaches $7.60\times$ on average and exceeds $10\times$ on multilingual and coding content at $B=32$, and grows nearly linearly with $B$, from $2.89\times$ at $B=4$ to $7.60\times$ at $B=32$.
(2) Confidence-based sampling, by contrast, achieves only $\sim$3$\times$ TPF at comparable accuracy in the block-$32$ diffusion-mode results of Sec.~\ref{sec:exp_instruct}, leaving a notable gap to the $7.60\times$ SOL ceiling. This indicates that confidence-based sampling is far from optimal and that there is substantial headroom for better-designed samplers to capture.
(3) Per-category SOL spans a $\sim$3.2$\times$ range, from $3.49\times$ on roleplay to $11.26\times$ on multilingual content. We hypothesize this reflects token-level entropy: templated content has more positions that are confidently determined by partial context, while open-ended generation does not. If this holds, samplers that adapt to the local content type could exploit this variance.
(4) A moderately small block length achieves the best accuracy, while larger block lengths degrade it. The trade-off between benchmark accuracy and efficiency, i.e., acceptance rate, indicates that improving diffusion performance under larger block lengths is a critical direction.

\begin{table}[t]
\centering
\caption{Per-category SOL acceptance ratio on SPEED-Bench under recursive dynamic compaction. Benchmark accuracy is averaged over 10 instruct LM benchmarks in Sec.~\ref{sec:exp_instruct}.
}
\vspace{-0.5em}
\label{tab:recd1-speedbench}
\resizebox{\columnwidth}{!}{
\begin{tabular}{lrrrr}
\toprule
Category & BL=32 & BL=16 & BL=8 & BL=4 \\
\midrule
coding         & 10.24 & 7.50 & 5.32 & 3.32 \\
humanities     &  6.93 & 5.00 & 3.76 & 2.76 \\
math           &  9.30 & 7.02 & 4.90 & 3.20 \\
multilingual   & 11.26 & 8.08 & 5.46 & 3.37 \\
qa             &  5.63 & 4.43 & 3.46 & 2.61 \\
rag            &  7.32 & 5.67 & 4.14 & 2.91 \\
reasoning      &  7.22 & 5.06 & 3.87 & 2.79 \\
roleplay       &  3.49 & 2.76 & 2.41 & 2.00 \\
stem           &  8.01 & 5.68 & 4.11 & 2.98 \\
summarization  &  6.02 & 4.48 & 3.25 & 2.71 \\
writing        &  6.13 & 4.71 & 3.58 & 2.62 \\
\midrule
\textbf{Acceptance rate} & \textbf{7.60} & \textbf{5.68} & \textbf{4.17} & \textbf{2.89} \\
\midrule
\textbf{Benchmark Acc} & \textbf{61.81} & \textbf{63.18} & \textbf{65.43} & \textbf{64.04} \\
\bottomrule
\end{tabular}
}
\vspace{-1.5em}
\end{table}

\noindent \textbf{Diffusion SOL vs. linear self-speculation.}
We additionally compare the SOL ceiling with linear self-speculation (Sec.~\ref{sec:tri-mode}) on SPEED-Bench at $B{=}32$, visualized in Fig.~\ref{fig:radar_sol_ss}. Two distinct metrics matter here: The \emph{acceptance rate} counts how many tokens are committed per acceptance step. SOL commits multiple positions in a single diffusion forward pass, while linear self-speculation accepts up to $k$ draft tokens per draft+verify cycle. The \emph{real TPF}, in contrast, is the average number of tokens committed per single model forward pass: SOL incurs only a per-block KV-cache recompute on top of one forward per acceptance step, so its real TPF is close to its acceptance rate; linear self-speculation, however, uses two forwards per cycle (one diffusion draft, one AR verification), so its real TPF is the acceptance rate divided by two. The two settings also target different correctness signals: SOL agrees with the diffusion mode's own serial-denoising output, while linear self-speculation agrees with the AR mode verification. We view this as a fair head-to-head measurement of parallel-decoding potential, considering that the diffusion and AR modes achieve comparable accuracy in Sec.~\ref{sec:exp_instruct} and thus both targets are equally credible references for a \emph{correct} token.

From Fig.~\ref{fig:radar_sol_ss}, we observe that:
(1) At the acceptance-rate level, linear self-speculation approaches SOL, achieving $6.82\times$ vs.\ $7.60\times$ overall (approximately $10.3\%$ below the upper bound), with similarly small gaps across categories. Thus, on top of diffusion drafting, applying AR verification is an effective way to approach the SOL ceiling.
(2) The real TPF gap, however, is much larger: $6.02\times$ for SOL vs.\ $3.41\times$ for linear self-speculation, i.e., a 76.5\% improvement. Beyond the doubled forward-pass cost, linear self-speculation only commits a contiguous prefix of the draft, discarding confident tokens beyond the first rejection. These two effects together motivate stronger diffusion-mode samplers that can safely commit tokens within a single forward pass, including at non-prefix positions.

\noindent \textbf{Implications for the tri-mode framework.}
The SOL analysis highlights two key takeaways for the tri-mode framework. \emph{(1)} Diffusion-mode decoding can be a promising approach for highly parallel decoding, provided that an optimized sampler can close the gap between pure confidence-based sampling and the SOL ceiling. \emph{(2)} AR verification is an effective way to sample from diffusion drafts and can approach SOL. However, its additional verification cost and prefix-only acceptance pattern fundamentally cap its real TPF below SOL, even when the diffusion drafter and the AR verifier are well aligned.

\vspace{-0.5em}
\section{\METHOD{} Family}
\label{sec:model_family}

We deliver the \METHOD{} model family in 3B, 8B, and 14B sizes, including base and instruct models as well as VLMs.

\vspace{-0.5em}
\subsection{Base Models}
\label{sec:base_model_setting}

To speed up pretraining, we start from the pretrained Ministral3~\cite{liu2026ministral} base models and apply the two-stage training strategy introduced in Sec.~\ref{sec:objectives}. Specifically, we adopt the pretraining dataset in~\cite{basant2025nvidia} and perform continuous pretraining for 1T tokens in Stage~1 (pure AR) and 300B tokens in Stage~2 (joint AR and diffusion training with an $\alpha$ of 0.3). The initial learning rate is set to $1\text{e-}5$ and decayed to $3\text{e-}6$ using a WSD schedule~\cite{hu2024minicpm} with the AdamW optimizer and a weight decay of 0.1. We adopt a global batch size of 512 and a sequence length of 4096. The training is performed on 256 NVIDIA H100 GPUs. 
We release the training and inference pipeline through \href{https://github.com/NVIDIA-NeMo/Megatron-Bridge/pull/3105}{Megatron Bridge}.

\vspace{-0.5em}
\subsection{Instruct Models}
\label{sec:instruct_model_setting}

We perform supervised fine-tuning (SFT) on top of our base models to deliver instruct models. Specifically, we adopt joint AR and diffusion training with an $\alpha$ of 0.3 throughout the SFT process. The initial learning rate is set to $2.5\text{e-}6$ and decayed to $2.5\text{e-}7$ using the WSD schedule~\cite{hu2024minicpm} with the AdamW optimizer and a weight decay of 0.1. We train the model on 45B tokens from the SFT dataset of~\cite{chandiramani2026nemotron}, with a global batch size of 256 and a sequence length of 16k. Following~\cite{nie2025large}, the training pipeline is the same as pretraining except that we do not mask any tokens from the prompt, and the loss is computed only on the answer parts. The training is performed on 256 NVIDIA H100 GPUs.

\begin{table*}[t]
\centering
\caption{
Benchmark our \METHOD{}-8B instruct model against SOTA AR and diffusion instruct LMs across scientific QA, instruction following, coding, and math reasoning benchmarks.
}
\vspace{-0.5em}
\resizebox{\textwidth}{!}{
\begin{tabular}{cccccccc>{\columncolor{nldgreen}}c>{\columncolor{nldgreen}}c>{\columncolor{nldgreen}}c>{\columncolor{nldgreen}}c}
\toprule
\textbf{Model}
& \begin{tabular}[c]{@{}c@{}}\textbf{Qwen2.5}\\ \textbf{7B}\end{tabular}
& \begin{tabular}[c]{@{}c@{}}\textbf{Qwen3}\\ \textbf{8B}\end{tabular}
& \begin{tabular}[c]{@{}c@{}}\textbf{Ministral3-8B}\\ \textbf{Instruct-2512}\end{tabular}
& \begin{tabular}[c]{@{}c@{}}\textbf{LLaDA-8B}\\ \textbf{Instruct}\end{tabular}
& \begin{tabular}[c]{@{}c@{}}\textbf{Dream-7B}\\ \textbf{Instruct}\end{tabular}
& \multicolumn{2}{c}{\begin{tabular}[c]{@{}c@{}}\textbf{SDAR-8B}\\ \textbf{Chat}\end{tabular}}
& \multicolumn{4}{>{\columncolor{nldgreen}}c}{\begin{tabular}[c]{@{}c@{}} \textbf{\METHOD{}-8B} \\ \textbf{(Tri-Mode in One Model)} \end{tabular}} \\
\midrule
Gen. Mode
& AR & AR & AR & Diff. & Diff. & Diff. & Diff. & AR & Diff. & Linear SS & Quad. SS \\
\midrule
\multicolumn{12}{c}{\textit{Scientific QA \& Instruction Following}} \\ \midrule
GPQA
& 37.12 & 49.24 & 42.87 & 33.30 & 33.00 & 40.20 & 30.80 & 44.44 & 43.94 & 40.40 & 44.30 \\
IFEval
& 74.58 & 87.38 & 64.31 & 59.90 & 62.50 & 61.40 & 60.07 & 68.65 & 68.32 & 69.13 & 71.00 \\
MMLU
& 74.86 & 76.66 & 73.90 & 65.50 & 67.00 & 78.60 & 78.83 & 79.85 & 78.71 & 79.01 & 79.95 \\
\midrule
\multicolumn{12}{c}{\textit{Coding}} \\ \midrule
HumanEval
& 77.44 & 81.71 & 71.04 & 49.40 & 55.50 & 78.70 & 79.27 & 80.49 & 78.66 & 81.71 & 79.27 \\
MBPP
& 81.55 & 81.88 & 78.97 & 41.00 & 58.80 & 72.00 & 67.32 & 85.19 & 83.86 & 84.92 & 85.19 \\
LCB-CPP
& 12.33 & 21.09 & 20.76 & 4.19 & 1.25 & 13.44 & 11.89 & 28.85 & 26.16 & 24.89 & 27.70 \\
\midrule
\multicolumn{12}{c}{\textit{Math}} \\ \midrule
Math500
& 75.10 & 84.80 & 83.60 & 39.20 & 43.00 & 78.60 & 72.40 & 88.00 & 85.80 & 87.60 & 88.80 \\
GSM8K
& 91.89 & 92.42 & 92.42 & 79.91 & 81.00 & 91.30 & 88.48 & 94.01 & 93.03 & 93.78 & 94.16 \\
AIME24
& 13.75 & 30.21 & 27.71 & 0.00 & 0.00 & 16.67 & 13.33 & 33.33 & 46.67 & 36.67 & 33.33 \\
AIME25
& 6.88 & 22.08 & 24.58 & 0.00 & 3.33 & 10.00 & 3.33 & 33.33 & 26.67 & 30.00 & 36.67 \\
\midrule
\multicolumn{12}{c}{\textit{Average over All Tasks}} \\ \midrule
Accuracy
& 54.55 & 62.75 & 58.02 & 37.24 & 40.54 & 54.09 & 50.57 & 63.61 & 63.18 & 62.81 & 64.04 \\
TPF
& 1.00 & 1.00 & 1.00 & 1.00 & 1.00 & 1.00 & 1.75 & 1.00 & 2.57 & 5.99 & 6.38 \\
\bottomrule
\end{tabular}
}
\label{tab:benchmark_instruct}
\vspace{-1em}
\end{table*}

\vspace{-0.5em}
\subsection{Vision-Language Models}

We extend \METHOD{} to the vision-language setting by adding a vision encoder and a multimodal projector to the diffusion LM backbone. The resulting VLM inherits the joint AR-diffusion training objective, the dual-stream attention pattern, and the tri-mode inference capability.  Below, we describe how the VLM is initialized from pretrained components and how image features are integrated into the dual-stream training layout.

\noindent \textbf{Architecture.} The VLM augments \METHOD{} with a vision encoder and a two-layer MLP projector with $2\times 2$ patch merging. The diffusion training and inference pipeline is shared with the text-only model, with only the vision frontend added.                                       

\noindent \textbf{Weight initialization.}  We initialize the LM backbone and LM head from the \METHOD{}-8B instruct model, which carries the diffusion-aware representations learned during joint AR-diffusion training, and initialize the vision encoder and projector from the corresponding AR VLM (\texttt{Ministral3-8B-Instruct-2512}) from the same model family, which provides fully trained visual perception and cross-modal alignment weights.

Because the LM architectures are identical between the two sources, the merge is exact with no parameter mismatch or interpolation. No new parameters are introduced; the vocabulary and embedding dimensions remain unchanged.

\noindent \textbf{Continued SFT.} Starting from the merged initialization, we finetune the full model (LM backbone, vision encoder, and projector) with the same joint AR-diffusion objective used for the text-only instruct model, on multimodal instruction-following data~\cite{wiedmann2025finevisionopendataneed}. 

\noindent  \textbf{Asymmetric dual stream.} A straightforward extension doubles all tokens in the noisy half, including vision tokens. However, vision tokens are never masked; only text response tokens are subject to forward corruption.  Carrying vision tokens in the noisy half, therefore, adds FLOPs without contributing to the diffusion loss. For high-resolution images, this overhead is substantial. We address this with an \textit{asymmetric dual-stream} layout that strips all vision token positions from the noisy half:
  \begin{equation}
    \bigl[\;\tilde{x}_t^{(\text{text},\; L_{\text{text}})}\;\mid\; 
    x^{(L)}\;\bigr],
    \qquad L_{\text{text}} = L - N_{\text{vis}},
    \label{eq:asymmetric}
  \end{equation}
where $N_{\text{vis}}$ is the number of vision tokens. The clean half retains the full sequence, including vision tokens, preserving complete visual context for the AR objective and for cross-stream conditioning.
The total sequence length becomes $L_{\text{text}} + L$ instead of $2L$, and the reduction of FLOPs of attention scales with the vision tokens $N_{\text{vis}} / L$.

\section{Evaluation and Analysis}
\label{sec:exp}

\subsection{Benchmark Instruct Models}
\label{sec:exp_instruct}

\noindent \textbf{Baselines and benchmarks.}
We compare our \METHOD{}-8B instruct model against SOTA AR instruct models (Qwen3-8B, Qwen2.5-7B, and Ministral3-8B Instruct) and SOTA diffusion instruct models (LLaDA-8B Instruct~\cite{nie2025large}, Dream-7B Instruct~\cite{ye2025dream}, and SDAR-8B Chat~\cite{cheng2025sdar}). We evaluate all modes of our model: AR, diffusion, and self-speculation, including both linear and quadratic self-speculation modes (denoted as linear SS and quadratic SS). We use LoRA-enhanced linear self-speculation by default.
For the diffusion model evaluation of \METHOD{}-8B and SDAR-8B Chat~\cite{cheng2025sdar}, we also report tokens per forward (TPF) by selecting different denoising thresholds~\cite{wu2025fast}. All models are evaluated in the non-thinking mode.

We evaluate across scientific QA and instruction following (GPQA, IFEval, MMLU), coding (HumanEval, MBPP, LiveCodeBench-CPP), and math reasoning (Math500, GSM8K, AIME24, AIME25). We use NeMo-Skills~\cite{nvidia_nemo_skills_2024} as the evaluation framework for all AR baselines and our \METHOD{}-8B, and use the official evaluation pipelines provided in the original papers for the diffusion baselines~\cite{nie2025large,ye2025dream,cheng2025sdar}.

\begin{table*}[t]
\centering
\caption{Per-task TPF achieved by linear self-speculation w/ and w/o LoRA tuning across 3B/8B/14B scales.}
\vspace{-0.5em}
\label{tab:lora_self_speculation}
\resizebox{\textwidth}{!}{
\begin{tabular}{cccccccccccccc}
\toprule
\textbf{Model Scale} & \textbf{Setting} & \textbf{GPQA} & \textbf{IFEval} & \textbf{HumanEval} & \textbf{MBPP} & \textbf{Math500} & \textbf{GSM8K} & \textbf{AIME24} & \textbf{AIME25} & \textbf{MMLU} & \textbf{LCB-CPP} & \textbf{Avg TPF} & \textbf{Avg Acc} \\
\midrule
\multirow{2}{*}{3B}
& w/o LoRA & 3.07 & 2.97 & 4.77 & 3.74 & 4.94 & 4.01 & 4.33 & 4.53 & 2.42 & 3.34 & 3.81 & 55.00 \\
& w/ LoRA  & 3.57 & 3.30 & 5.56 & 4.23 & 5.63 & 4.55 & 4.99 & 5.11 & 2.74 & 3.95 & 4.36 & 55.00 \\
\midrule
\multirow{2}{*}{8B}
& w/o LoRA & 5.10 & 4.32 & 4.62 & 3.44 & 5.43 & 4.47 & 5.38 & 5.05 & 3.25 & 4.14 & 4.52 & 62.88 \\
& w/ LoRA  & 6.64 & 5.52 & 5.82 & 4.44 & 7.36 & 5.89 & 7.44 & 6.92 & 4.08 & 5.70 & 5.99 & 62.81 \\
\midrule
\multirow{2}{*}{14B}
& w/o LoRA & 5.31 & 3.86 & 6.75 & 4.42 & 5.01 & 4.54 & 4.47 & 3.92 & 4.95 & 3.42 & 4.67 & 66.35 \\
& w/ LoRA  & 6.07 & 4.74 & 8.11 & 5.22 & 6.72 & 5.79 & 5.88 & 5.41 & 7.22 & 4.46 & 5.96 & 66.36 \\
\bottomrule
\end{tabular}
}
\vspace{-0.5em}
\end{table*}

\noindent \textbf{Observations.}
As shown in Tab.~\ref{tab:benchmark_instruct}, we observe that, compared to SOTA AR and diffusion instruct LMs, our \METHOD{}-8B achieves both higher accuracy and efficiency across all modes. More specifically,
(1) In terms of AR performance, \METHOD{}-8B in AR mode delivers +0.86\% higher average accuracy than Qwen3-8B and outperforms all other AR baselines, demonstrating that the joint AR–diffusion training objective effectively preserves strong AR accuracy. In fact, the ablation study under a controlled setting in Sec.~\ref{sec:exp_mutual} indicates that adding the diffusion objective can maintain or slightly improve AR accuracy, potentially due to an improved ability to predict the future.
(2) The diffusion mode decodes $2.57\times$ TPF while achieving $+0.43\%$ higher average accuracy than Qwen3-8B. Compared to existing diffusion LMs, \METHOD{}-8B outperforms SDAR-8B Chat by $+9.09\%$ in average accuracy and better maintains accuracy under larger decoding parallelism, as shown in Fig.~\ref{fig:teaser} (b).
(3) LoRA-tuned linear self-speculation maintains comparable accuracy to the diffusion mode while further boosting TPF to 5.99$\times$, indicating the effectiveness of aligning the diffusion drafter with the AR target via lightweight LoRA tuning.
(4) Quadratic self-speculation can achieve the highest TPF of 6.38$\times$, as it prepares the next block for all possible acceptance positions at a quadratic cost. However, due to the use of FlexAttention with less optimized kernels for the dedicated attention mask~\cite{liu2025tidar}, the real-device efficiency of quadratic self-speculation falls behind the linear one according to Fig.~\ref{fig:teaser} (b). As such, we use linear self-speculation by default.

\noindent \textbf{Remark.} The tri-mode design enables \METHOD{} to serve different deployment needs within a single model:
(1) The AR mode matches or surpasses SOTA AR LMs in accuracy, meaning that \METHOD{} can serve as a drop-in replacement for any application that currently uses an AR model, with no pipeline changes required.
(2) The diffusion mode enables one-for-all flexibility: by adjusting the denoising threshold, a single model can achieve a range of accuracy-throughput trade-offs, as illustrated in Fig.~\ref{fig:teaser} (b). 
(3) Self-speculation is promising for achieving significant inference speedup through the synergy between AR and diffusion. While it sacrifices the flexibility of the diffusion mode by only accepting prefix tokens, it provides a reliable mechanism to verify diffusion drafts, which can lead to substantial inference acceleration, as demonstrated in Sec.~\ref{sec:exp_mtp}.

\begin{figure}[t]
\centering
\vspace{-1em}
\includegraphics[width=0.95\linewidth]{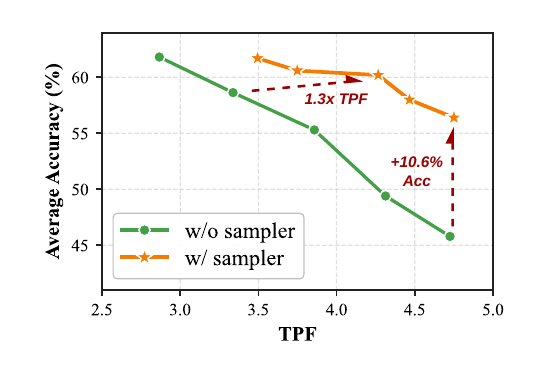} 
\vspace{-2em}
\caption{Comparing the accuracy-TPF trade-offs achieved w/ and w/o a sampler.
}
\label{fig:sampler}
\vspace{-1.5em}
\end{figure}

\begin{table*}[t]
\centering
\caption{
Benchmark our \METHOD{}-3B/14B instruct models against SOTA AR instruct models.
}
\vspace{-0.5em}
\label{tab:benchmark_3b_14b}
\resizebox{\textwidth}{!}{
\begin{tabular}{cccccccccccccc}
\toprule
\textbf{Model} & \textbf{Gen. Mode} & \textbf{GPQA} & \textbf{IFEval} & \textbf{HumanEval} & \textbf{MBPP} & \textbf{Math500} & \textbf{GSM8K} & \textbf{AIME24} & \textbf{AIME25} & \textbf{MMLU} & \textbf{LCB-CPP} & \textbf{Avg Acc} & \textbf{Avg TPF} \\
\midrule
\multicolumn{14}{c}{\textit{3B Scale}} \\ \midrule
Llama-3.2-3B-Instruct        & AR & 27.78 & 69.69 & 65.85 & 66.93 & 32.60 & 62.09 & 0.00  & 0.00  & 62.45 & 9.25  & 39.90 & 1.00 \\
Phi-4-mini-Instruct          & AR & 40.91 & 71.16 & 77.44 & 75.13 & 70.80 & 76.57 & 10.00 & 6.67  & 53.37 & 9.91  & 49.20 & 1.00 \\
Ministral3-3B-Instruct-2512  & AR & 28.79 & 57.55 & 64.79 & 68.39 & 71.90 & 87.41 & 12.50 & 12.71 & 63.21 & 12.50 & 47.97 & 1.00 \\
Qwen3-4B                     & AR & 37.18 & 72.20 & 75.91 & 72.49 & 68.40 & 92.19 & 10.63 & 10.63 & 77.05 & 15.58 & 53.23 & 1.00 \\
\midrule
\cellcolor{nldgreen} & \cellcolor{nldgreen}AR & \cellcolor{nldgreen}39.39 & \cellcolor{nldgreen}69.39 & \cellcolor{nldgreen}76.22 & \cellcolor{nldgreen}71.16 & \cellcolor{nldgreen}77.60 & \cellcolor{nldgreen}87.87 & \cellcolor{nldgreen}23.33 & \cellcolor{nldgreen}16.67 & \cellcolor{nldgreen}71.70 & \cellcolor{nldgreen}21.37 & \cellcolor{nldgreen}55.50 & \cellcolor{nldgreen}1.00 \\
\cellcolor{nldgreen} & \cellcolor{nldgreen}Diff. & \cellcolor{nldgreen}33.84 & \cellcolor{nldgreen}68.93 & \cellcolor{nldgreen}74.39 & \cellcolor{nldgreen}73.54 & \cellcolor{nldgreen}74.80 & \cellcolor{nldgreen}88.40 & \cellcolor{nldgreen}16.67 & \cellcolor{nldgreen}10.00 & \cellcolor{nldgreen}72.06 & \cellcolor{nldgreen}16.30 & \cellcolor{nldgreen}52.90 & \cellcolor{nldgreen}1.91 \\
\cellcolor{nldgreen} & \cellcolor{nldgreen}Linear SS & \cellcolor{nldgreen}35.86 & \cellcolor{nldgreen}68.96 & \cellcolor{nldgreen}75.00 & \cellcolor{nldgreen}70.11 & \cellcolor{nldgreen}77.40 & \cellcolor{nldgreen}87.79 & \cellcolor{nldgreen}26.67 & \cellcolor{nldgreen}16.67 & \cellcolor{nldgreen}71.89 & \cellcolor{nldgreen}20.04 & \cellcolor{nldgreen}55.00 & \cellcolor{nldgreen}4.36 \\
\cellcolor{nldgreen}\multirow{-4}{*}{\makecell{\textbf{Nemotron-Labs}\\\textbf{-Diffusion-3B}\\\textbf{(Tri-Mode)}}} & \cellcolor{nldgreen}Quad.\ SS & \cellcolor{nldgreen}42.93 & \cellcolor{nldgreen}71.36 & \cellcolor{nldgreen}79.27 & \cellcolor{nldgreen}76.46 & \cellcolor{nldgreen}78.20 & \cellcolor{nldgreen}88.25 & \cellcolor{nldgreen}13.33 & \cellcolor{nldgreen}16.67 & \cellcolor{nldgreen}72.12 & \cellcolor{nldgreen}19.60 & \cellcolor{nldgreen}55.80 & \cellcolor{nldgreen}5.42 \\
\midrule
\multicolumn{14}{c}{\textit{14B Scale}} \\ \midrule
Gemma-3-12B-IT               & AR & 38.32 & 85.73 & 58.23 & 85.45 & 84.55 & 90.45 & 23.75 & 16.88 & 76.41 & 20.54 & 58.03 & 1.00 \\
Phi-3-Medium-14B             & AR & 37.56 & 85.75 & 70.43 & 76.65 & 43.35 & 89.69 & 1.88  & 0.62  & 76.67 & 10.79 & 49.34 & 1.00 \\
Phi-4-14B                    & AR & 56.94 & 68.96 & 84.60 & 83.93 & 79.95 & 92.27 & 19.17 & 15.83 & 84.76 & 21.75 & 60.82 & 1.00 \\
Ministral3-14B-Instruct-2512 & AR & 52.02 & 71.51 & 72.56 & 82.47 & 86.30 & 92.80 & 36.25 & 29.38 & 79.88 & 26.54 & 62.97 & 1.00 \\
Qwen3-14B                    & AR & 50.51 & 88.36 & 83.54 & 87.30 & 85.40 & 94.31 & 33.33 & 20.00 & 81.51 & 27.42 & 65.17 & 1.00 \\
\midrule
\cellcolor{nldgreen} & \cellcolor{nldgreen}AR & \cellcolor{nldgreen}54.55 & \cellcolor{nldgreen}68.50 & \cellcolor{nldgreen}86.59 & \cellcolor{nldgreen}85.19 & \cellcolor{nldgreen}88.40 & \cellcolor{nldgreen}91.36 & \cellcolor{nldgreen}46.67 & \cellcolor{nldgreen}43.33 & \cellcolor{nldgreen}82.51 & \cellcolor{nldgreen}27.48 & \cellcolor{nldgreen}67.46 & \cellcolor{nldgreen}1.00 \\
\cellcolor{nldgreen} & \cellcolor{nldgreen}Diff. & \cellcolor{nldgreen}48.99 & \cellcolor{nldgreen}69.03 & \cellcolor{nldgreen}83.54 & \cellcolor{nldgreen}82.80 & \cellcolor{nldgreen}85.80 & \cellcolor{nldgreen}93.71 & \cellcolor{nldgreen}43.33 & \cellcolor{nldgreen}50.00 & \cellcolor{nldgreen}82.17 & \cellcolor{nldgreen}25.77 & \cellcolor{nldgreen}66.51 & \cellcolor{nldgreen}2.74 \\
\cellcolor{nldgreen} & \cellcolor{nldgreen}Linear SS & \cellcolor{nldgreen}47.47 & \cellcolor{nldgreen}70.06 & \cellcolor{nldgreen}85.37 & \cellcolor{nldgreen}84.66 & \cellcolor{nldgreen}86.60 & \cellcolor{nldgreen}92.04 & \cellcolor{nldgreen}50.00 & \cellcolor{nldgreen}40.00 & \cellcolor{nldgreen}81.11 & \cellcolor{nldgreen}26.32 & \cellcolor{nldgreen}66.36 & \cellcolor{nldgreen}5.96 \\
\cellcolor{nldgreen}\multirow{-4}{*}{\makecell{\textbf{Nemotron-Labs}\\\textbf{-Diffusion-14B}\\\textbf{(Tri-Mode)}}} & \cellcolor{nldgreen}Quad.\ SS & \cellcolor{nldgreen}52.02 & \cellcolor{nldgreen}72.15 & \cellcolor{nldgreen}87.20 & \cellcolor{nldgreen}85.45 & \cellcolor{nldgreen}88.00 & \cellcolor{nldgreen}92.12 & \cellcolor{nldgreen}53.33 & \cellcolor{nldgreen}40.00 & \cellcolor{nldgreen}82.45 & \cellcolor{nldgreen}28.74 & \cellcolor{nldgreen}68.15 & \cellcolor{nldgreen}6.92 \\
\bottomrule
\end{tabular}
}
\end{table*}

\begin{table*}[t]
\centering
\caption{
Benchmark our \METHOD{}-8B base model against SOTA AR and diffusion base LMs across coding, math, knowledge, and commonsense reasoning benchmarks.
}
\vspace{-0.5em}
\label{tab:benchmark_base}
\resizebox{\textwidth}{!}{
\begin{tabular}{cccccccccccccccc}
\toprule
\textbf{Model} & \textbf{Gen. Mode} & \makecell{\textbf{Human}\\\textbf{Eval}} & \makecell{\textbf{Human}\\\textbf{Eval+}} & \textbf{MBPP} & \textbf{MBPP+} & \textbf{GSM8K} & \makecell{\textbf{Minerva}\\\textbf{Math}} & \textbf{MMLU} & \textbf{ARC-E} & \textbf{ARC-C} & \makecell{\textbf{Hella}\\\textbf{swag}} & \textbf{PIQA} & \makecell{\textbf{Wino}\\\textbf{grande}} & \makecell{\textbf{Avg}\\\textbf{Acc}} & \makecell{\textbf{Avg}\\\textbf{TPF}} \\
\midrule
Llama-3.1-8B  & AR    & 35.37 & 28.66 & 48.80 & 61.90 & 54.06 & 18.22 & 65.15 & 81.31 & 53.41 & 78.93 & 81.18 & 77.43 & 57.04 & 1.00 \\
Ministral3-8B & AR    & 42.68 & 38.41 & 61.60 & 76.98 & 80.21 & 44.58 & 76.39 & 86.15 & 60.75 & 79.01 & 80.74 & 73.48 & 66.75 & 1.00 \\
Qwen3-8B      & AR    & 64.63 & 56.71 & 69.40 & 83.07 & 86.73 & 52.94 & 76.93 & 81.90 & 53.16 & 78.59 & 79.22 & 75.69 & 71.58 & 1.00 \\
LLaDA-8B      & Diff. & 32.32 & 27.44 & 40.80 & 51.85 & 70.96 & 27.30 & 65.86 & 73.78 & 49.15 & 71.05 & 73.88 & 74.66 & 54.92 & 1.00 \\
Dream-7B      & Diff. & 54.88 & 49.39 & 56.80 & 74.60 & 77.18 & 39.60 & 67.00 & 82.20 & 59.13 & 73.73 & 75.52 & 73.56 & 65.30 & 1.00 \\
\midrule
\cellcolor{nldgreen} & \cellcolor{nldgreen}AR & \cellcolor{nldgreen}60.37 & \cellcolor{nldgreen}53.05 & \cellcolor{nldgreen}68.20 & \cellcolor{nldgreen}82.54 & \cellcolor{nldgreen}88.25 & \cellcolor{nldgreen}66.00 & \cellcolor{nldgreen}74.68 & \cellcolor{nldgreen}83.38 & \cellcolor{nldgreen}58.11 & \cellcolor{nldgreen}76.08 & \cellcolor{nldgreen}80.09 & \cellcolor{nldgreen}71.98 & \cellcolor{nldgreen}71.89 & \cellcolor{nldgreen}1.00 \\
\cellcolor{nldgreen} & \cellcolor{nldgreen}Diff. & \cellcolor{nldgreen}62.80 & \cellcolor{nldgreen}57.32 & \cellcolor{nldgreen}67.00 & \cellcolor{nldgreen}81.75 & \cellcolor{nldgreen}87.26 & \cellcolor{nldgreen}65.16 & \cellcolor{nldgreen}74.68 & \cellcolor{nldgreen}83.38 & \cellcolor{nldgreen}58.11 & \cellcolor{nldgreen}76.08 & \cellcolor{nldgreen}80.09 & \cellcolor{nldgreen}71.98 & \cellcolor{nldgreen}72.13 & \cellcolor{nldgreen}2.06 \\
\cellcolor{nldgreen} & \cellcolor{nldgreen}Linear SS & \cellcolor{nldgreen}63.41 & \cellcolor{nldgreen}56.10 & \cellcolor{nldgreen}67.20 & \cellcolor{nldgreen}81.75 & \cellcolor{nldgreen}88.17 & \cellcolor{nldgreen}67.38 & \cellcolor{nldgreen}74.68 & \cellcolor{nldgreen}83.38 & \cellcolor{nldgreen}58.11 & \cellcolor{nldgreen}76.08 & \cellcolor{nldgreen}80.09 & \cellcolor{nldgreen}71.98 & \cellcolor{nldgreen}72.36 & \cellcolor{nldgreen}4.67 \\
\cellcolor{nldgreen}\multirow{-4}{*}{\makecell{\textbf{Nemotron-Labs}\\\textbf{-Diffusion-8B}\\\textbf{(Tri-Mode)}}} & \cellcolor{nldgreen}Quad. SS & \cellcolor{nldgreen}62.20 & \cellcolor{nldgreen}54.88 & \cellcolor{nldgreen}67.60 & \cellcolor{nldgreen}81.48 & \cellcolor{nldgreen}88.48 & \cellcolor{nldgreen}66.24 & \cellcolor{nldgreen}74.68 & \cellcolor{nldgreen}83.38 & \cellcolor{nldgreen}58.11 & \cellcolor{nldgreen}76.08 & \cellcolor{nldgreen}80.09 & \cellcolor{nldgreen}71.98 & \cellcolor{nldgreen}72.10 & \cellcolor{nldgreen}7.04 \\
\bottomrule
\end{tabular}
}
\vspace{-0.5em}
\end{table*}

\begin{table*}[t]
  \centering
  \caption{%
    Benchmarking discrete diffusion VLMs and \METHOD{}-VLM across tasks. The diffusion mode of \METHOD{}-VLM uses denoising threshold $\tau{=}0.9$.
  }
  \vspace{-0.5em}
  \label{tab:nemotron_vlm_expr}
  \resizebox{\textwidth}{!}{%
    \begin{tabular}{c c *{8}{c} c c}
      \toprule
      Model & Gen. Mode
        & AI2D & ChartQA & DocVQA & MMMU
        & \makecell{MMMU\\Pro-10c} & \makecell{MMMU\\Pro-V-CoT} & \makecell{Math\\Vista} & \makecell{RealWorld\\QA}
        & TPF & Acc \\
      \midrule
      MMaDA       & Diff.
        & 67.4 & 9.6 & 9.5 & 30.2
        & 16.5 & 8.5 & 33.4 & 49.2
        & 1 & 28.0 \\
      LaViDa      & Diff.
        & 70.0 & 59.0 & 64.6 & 43.3
        & 28.7 & 10.5 & 44.8 & 54.5
        & 1 & 46.9 \\
      Dimple      & Diff.
        & 74.4 & 63.4 & 37.7 & 45.2
        & 23.8 & 12.4 & 42.3 & 55.4
        & 1 & 44.3 \\
      LLaDA-V-8B  & Diff.
        & \textbf{77.8} & 78.3 & 83.9 & 48.6
        & \textbf{35.2} & 18.6 & 59.7 & \textbf{63.2}
        & 1 & 58.2 \\
      \midrule
      \cellcolor{nldgreen} & \cellcolor{nldgreen}AR & \cellcolor{nldgreen}75.0 & \cellcolor{nldgreen}\textbf{81.3} & \cellcolor{nldgreen}\textbf{89.2} & \cellcolor{nldgreen}50.3 & \cellcolor{nldgreen}32.6 & \cellcolor{nldgreen}\textbf{24.3} & \cellcolor{nldgreen}\textbf{60.4} & \cellcolor{nldgreen}62.6 & \cellcolor{nldgreen}1 & \cellcolor{nldgreen}\textbf{59.5} \\
      \cellcolor{nldgreen} & \cellcolor{nldgreen}Diff. & \cellcolor{nldgreen}74.7 & \cellcolor{nldgreen}76.6 & \cellcolor{nldgreen}88.3 & \cellcolor{nldgreen}\textbf{50.4} & \cellcolor{nldgreen}31.7 & \cellcolor{nldgreen}22.2 & \cellcolor{nldgreen}58.5 & \cellcolor{nldgreen}60.3 & \cellcolor{nldgreen}\makecell{2.46 all samples\\2.80 tok${>}$100\\3.15 tok${>}$200} & \cellcolor{nldgreen}57.9 \\ 
      \cellcolor{nldgreen}\multirow{-5}{*}{\makecell{\textbf{Nemotron-Labs}\\\textbf{-Diffusion}\\\textbf{-VLM-8B}}} & \cellcolor{nldgreen}Linear SS & \cellcolor{nldgreen}74.9 & \cellcolor{nldgreen}81.2 & \cellcolor{nldgreen}89.3 & \cellcolor{nldgreen}50.0 & \cellcolor{nldgreen}32.8 & \cellcolor{nldgreen}24.1 & \cellcolor{nldgreen}60.7 & \cellcolor{nldgreen}62.4 & \cellcolor{nldgreen}\makecell{3.63 all samples\\6.03 tok${>}$100\\7.45 tok${>}$200} & \cellcolor{nldgreen}59.4 \\
      \bottomrule
    \end{tabular}%
  }
\vspace{-0.5em}
\end{table*}

\noindent \textbf{Improve diffusion decoding with a better sampler.} We evaluate the effectiveness of the proposed sampler in Sec.~\ref{sec:diff_mode_inference}. We apply it on top of our instruct model with a block size of 32 and report the average accuracy across all ten tasks under different denoising thresholds to obtain the accuracy–TPF trade-off. As shown in Fig.~\ref{fig:sampler}, the trained sampler shifts the entire Pareto frontier upward, delivering higher TPF at the same accuracy (e.g., 1.3$\times$ TPF) or higher accuracy at the same TPF (e.g., +10.6\% accuracy). The improvements from this simple design suggest that part of the gap between the realized diffusion-mode TPF and the SOL ceiling in Sec.~\ref{sec:sol} can be closed by learning the acceptance policy itself.

\noindent \textbf{The impact of LoRA tuning for linear self-speculation.}
We perform an ablation study on the impact of LoRA tuning for aligning diffusion drafters and AR verifiers. As shown in Tab.~\ref{tab:lora_self_speculation}, we observe that (1) even without LoRA adapters, linear self-speculation already achieves nontrivial TPF, e.g., $4.67\times$ for our 14B model, with larger model scales generally leading to higher TPF; and (2) adding LoRA tuning consistently improves TPF, yielding $14.4\%/32.5\%/27.6\%$ relative gains at the 3B/8B/14B scales. We also note that the small accuracy gap between different self-speculation settings and the AR mode is due to kernel mismatches between 1-token decoding and multi-token prefilling.

\vspace{-0.5em}
\subsection{Extend to More Model Scales}
\label{sec:exp_3b_14b}

\noindent \textbf{Baselines and benchmarks.}
We extend the evaluation to two additional scales, \METHOD{}-3B/14B, under the same evaluation protocol as Sec.~\ref{sec:exp_instruct}, and benchmark against SOTA open-source AR instruct models at the corresponding scales.

\noindent \textbf{Observations.}
As shown in Tab.~\ref{tab:benchmark_3b_14b}, we observe that:
(1) \METHOD{} maintains consistent improvements in accuracy and efficiency across scales and generation modes. For example, using LoRA-tuned linear self-speculation, our \METHOD{}-3B/14B outperforms the strongest baselines, Qwen3-4B/14B, by +1.77\%/+1.19\% in accuracy while achieving 4.36$\times$/5.96$\times$ TPF, respectively. 
(2) Based on the performance of \METHOD{}-3B/8B/14B, larger LMs generally more readily unlock parallel diffusion abilities, as the TPF of diffusion/self-speculation modes grows broadly with scale. For example, the TPF of linear self-speculation increases from $4.36\times$ to $5.96\times$ when scaling from 3B to 14B. We attribute this to the stronger future-prediction abilities of larger models, which yield more reliable draft predictions.

\subsection{Benchmark Base Models}
\label{sec:exp_base}

\noindent \textbf{Baselines and benchmarks.}
We compare \METHOD{}-8B against SOTA AR base LMs and two representative diffusion LMs (LLaDA-8B~\cite{nie2025large} and Dream-7B~\cite{ye2025dream}).
We evaluate on coding benchmarks (HumanEval, HumanEval+, MBPP, MBPP+), math reasoning (GSM8K, Minerva Math), knowledge (MMLU), and commonsense reasoning (ARC-E, ARC-C, Hellaswag, PIQA, Winogrande). 

\noindent \textbf{Observations.}
As shown in Tab.~\ref{tab:benchmark_base}, we observe findings consistent with the instruct model results. Our \METHOD{}-8B base model achieves both higher accuracy and efficiency across all modes: (1) the AR mode delivers +5.14\%/+0.31\% higher average accuracy than Ministral3-8B/Qwen3-8B; (2) the diffusion mode delivers +17.21\%/+6.83\% higher accuracy than LLaDA-8B/Dream-7B; and (3) self-speculation achieves 4.67$\times$ TPF (linear) and 7.04$\times$ TPF (quadratic) with over 0.5\% higher average accuracy compared to the strongest baseline, Qwen3-8B.

\begin{figure*}[b!]
\centering
\vspace{-0.5em}
\includegraphics[width=\linewidth]{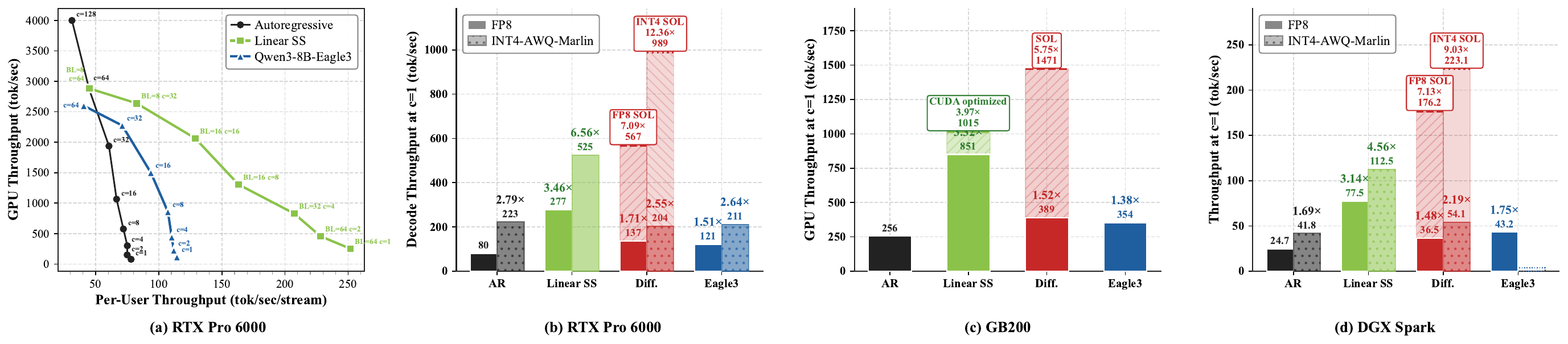} 
\vspace{-1.5em}
\caption{(a) System vs. per-user throughput trade-off on an NVIDIA RTX Pro 6000 GPU; (b)/(c)/(d): The throughput under a concurrency of 1 on NVIDIA RTX Pro 6000, GB200, and DGX Spark, respectively.}
\label{fig:inference_sglang}
\end{figure*}

\vspace{-0.5em}
\subsection{Benchmark VLMs}
\label{sec:exp_vlm}

\noindent \textbf{Benchmarks and evaluation settings.}
We evaluate on a diverse set of VLM benchmarks spanning two categories. Short-answer benchmarks require brief, factual responses: AI2D~\citep{kembhavi2016ai2d}, ChartQA~\citep{masry2022chartqa},  DocVQA~\citep{mathew2021docvqa}, MMMU~\citep{yue2024mmmu}, MathVista~\citep{lu2024mathvista}, and RealWorldQA~\citep{realworldqa2024}. Long-answer benchmarks require extended chain-of-thought reasoning: MMMU-Pro-V~\cite{yue2025mmmupro}. We benchmark against existing diffusion VLMs~\cite{yang2025mmada, li2025lavida, you2025llada, yu2025dimple}. All benchmarks are evaluated using VLMEvalKit~\citep{duan2024vlmevalkit} under the same prompts and post-processing as the AR baseline (Ministral3 VLM). Throughput (tokens per second, TPS) is measured on a single NVIDIA H100 GPU with identical prompt batching for fair comparison.

\noindent \textbf{Observations.}
As shown in Tab.~\ref{tab:nemotron_vlm_expr}, we compare our \METHOD{}-VLM against existing diffusion VLMs in three modes: diffusion, AR, and linear self-speculation decoding. We observe that
(1) In terms of AR performance, our model delivers 1.3\% higher average accuracy than the strongest baseline, LLaDA-V-8B.
(2) The diffusion mode provides 2.46$\times$–3.15$\times$ TPF while maintaining competitive accuracy.
(3) Linear self-speculation preserves near-AR accuracy, with only a 0.1\% average accuracy drop, while further increasing decoding parallelism to 3.63$\times$–7.45$\times$ TPF, where the higher end is achieved for responses exceeding 200 tokens. This implies that the advantage of our model is most pronounced on tasks requiring longer reasoning.
These results demonstrate that the joint AR–diffusion training framework extends effectively to the vision-language setting, preserving the broad capabilities of the LM backbone while enabling efficient multi-token decoding.

\subsection{Inference Efficiency}
\label{sec:exp_mtp}

We analyze and compare the deployment efficiency of \METHOD{} against MTP/Eagle3-style speculative decoding.

\noindent \textbf{Self-speculation vs. MTP.}
MTP methods such as Eagle3~\cite{li2025eagle} have become the default choice for efficient LLM deployment at low concurrency, where a small model is used to draft multiple future tokens, and then a single forward pass of a larger AR model verifies and accepts some of them. This schedule is more efficient at low concurrency because the memory transfer cost is similar between token-by-token generation and verification passes, while the latter can accept multiple tokens in a single pass. The two main bottlenecks of Eagle3 are: (1) the draft model has limited capacity and is less reliable beyond a short horizon; and (2) proposals are generated recursively, so even if the draft model is tiny, it still incurs the cost of the embedding layer and LM head. In contrast, \METHOD{} provides unique advantages through (1) significantly higher acceptance length, and (2) token-parallel drafting that enables better GPU utilization.

\noindent \textbf{Setup.}
We deploy \METHOD{}-8B with the SGLang server and profile it on NVIDIA GB200, RTX Pro 6000, and DGX Spark under different concurrency levels, comparing against Qwen3-8B-Eagle3, with results shown in Fig.~\ref{fig:teaser} (c) and Fig.~\ref{fig:inference_sglang}. Evaluations are conducted on SPEED-Bench~\cite{abramovich2026speed} across four categories (math, coding, reasoning, and multilingual), and limiting generation length to 1024 tokens to avoid repetition/hallucinations. We perform a grid search over hyperparameters for Eagle3, whereas for \METHOD{} we only vary the \textit{block length}. We additionally report a SOL throughput estimate mentioned in Sec.~\ref{sec:sol}, providing a reference ceiling for current self-speculation infrastructure.

\noindent \textbf{Observations.}
As shown in Fig.~\ref{fig:teaser} (c) and Fig.~\ref{fig:inference_sglang}, we observe that:
(1) Linear self-speculation consistently improves user throughput over the AR mode across all three GPUs, achieving up to $3.3\times$ speedup over AR on GB200 ($3.97\times$ speedup and 1015 tok/sec with an optimized kernel), as shown in Fig.~\ref{fig:inference_sglang} (c), and pushing the absolute throughput at batch size 1 to $277$/$525$ tok/sec on RTX Pro 6000 ($3.46\times$/$2.35\times$ over AR) and $77.5$/$112.5$ tok/sec on DGX Spark ($3.14\times$/$2.69\times$ over AR) under FP8/INT4 quantization, demonstrating its effectiveness as a drop-in low-concurrency acceleration scheme.
(2) Compared with Eagle3, linear self-speculation delivers a $2.4\times$/$2.3\times$/$1.8\times$ speedup at batch size 1 on GB200/RTX Pro 6000/DGX Spark and achieves better trade-offs between system throughput and per-user throughput, as shown in Fig.~\ref{fig:teaser} (c) and Fig.~\ref{fig:inference_sglang} (a). This indicates that diffusion drafting paired with AR verification is a more effective acceleration mechanism than auxiliary-head MTP due to its higher acceptance length.
(3) The SOL ceiling reveals substantial remaining headroom: on RTX Pro 6000, the projected SOL throughput reaches $7.09\times$/$12.36\times$ over AR under FP8/INT4 quantization, roughly $2\times$ above linear self-speculation.

\noindent \textbf{Acceptance length per category.}
As shown in Tab.~\ref{tab:accept_len_per_category}, \METHOD{} achieves significantly higher acceptance length than both Eagle3 and MTP across all categories, with average acceptance lengths of $5.46$/$6.82$ for Native/LoRA-tuned \METHOD{} versus $2.75$/$4.24$ for Eagle3/MTP. The gap further widens to $6.75$/$8.69$ vs.\ $2.81$/$4.73$ on the four diffusion-friendly categories (coding, math, reasoning, multilingual), implying that diffusion drafting yields more reliable multi-token proposals, especially on structured tasks with strong syntactic or semantic constraints.

\begin{table}[t]
  \centering
  \caption{Per-category acceptance length on SPEED-Bench~\cite{abramovich2026speed}.
  Comparing Native / LoRA for \METHOD{}-8B and Qwen3-8B-Eagle3 / Qwen3-9B-MTP, all with draft length 31.}
  \vspace{-0.5em}
  \label{tab:accept_len_per_category}
  \resizebox{\columnwidth}{!}{%
    \begin{tabular}{lrrrr}
      \toprule
      \textbf{Category} & \textbf{Native} & \textbf{LoRA} & \textbf{Eagle3} & \textbf{MTP} \\
      \midrule
      \rowcolor{gray!12}
      coding        & 6.61 & 8.57  & 3.14 & 5.97 \\
      \rowcolor{gray!12}
      math          & 6.24 & 8.14  & 2.79 & 4.80 \\
      \rowcolor{gray!12}
      reasoning     & 6.18 & 7.99  & 3.40 & 3.68 \\
      \rowcolor{gray!12}
      multilingual  & 7.96 & 10.06 & 1.91 & 4.47 \\
      humanities    & 5.01 & 6.31  & 3.12 & 3.76 \\
      qa            & 4.01 & 4.65  & 2.63 & 3.50 \\
      rag           & 5.07 & 6.15  & 3.06 & 4.75 \\
      roleplay      & 4.66 & 5.54  & 2.10 & 2.32 \\
      stem          & 5.55 & 7.02  & 2.92 & 4.45 \\
      summarization & 4.47 & 5.48  & 2.66 & 3.69 \\
      writing       & 4.28 & 5.07  & 2.81 & 3.21 \\
      \midrule
      \textbf{Average} & \textbf{5.46} & \textbf{6.82} & \textbf{2.75} & \textbf{4.24} \\
      \rowcolor{gray!18}
      \textbf{4 category avg} & \textbf{6.75} & \textbf{8.69} & \textbf{2.81} & \textbf{4.73} \\
      \bottomrule
    \end{tabular}%
  }
\vspace{-1em}
\end{table}
\vspace{-0.5em}
\section{Related Work}
\label{sec:related-work}

\textbf{Diffusion language models.}
To overcome the token-by-token decoding nature of AR LMs, diffusion LMs, both continuous~\cite{li2022diffusion,gong2022diffuseq,han2022ssd} and discrete~\cite{austin2021structured,he2022diffusionbert,sahoo2024simple,shi2024simplified,lou2023discrete,ou2024your}, have been proposed to perform non-AR decoding and thus enable parallel token generation. Among them, masked diffusion LMs~\cite{he2022diffusionbert,shi2024simplified,sahoo2024simple,nie2025large,ye2025dream} have been successfully scaled up (e.g., LLaDA~\cite{nie2025large} and Dream~\cite{ye2025dream}). Follow-up work has further explored alternative diffusion LM paradigms~\cite{sahoo2025esoteric,sahoo2025diffusion,xue2025any}, and scaled them to larger scales~\cite{bie2025llada2,bie2026llada2,gemini-diffusion} or domain-specific specialists such as coding agents~\cite{khanna2025mercury,song2025seed,gong2025diffucoder,xie2025dream}, explored dedicated reinforcement learning schemes~\cite{zhao2025d1,zhu2025llada}, and extended them to more modalities~\cite{you2025llada,yang2025mmada}. 
Compared to AR LMs, diffusion LMs have been demonstrated to be better learners under data-constrained settings~\cite{prabhudesai2025diffusion} and show improved performance in planning~\cite{ye2025dream} and text embedding~\cite{zhang2025diffusion}.

\textbf{Diffusion language model acceleration.} 
Despite the acceleration potential of large diffusion LMs~\cite{nie2025large,ye2025dream}, the gap between bidirectional attention and KV caching, along with the one-token-per-step denoising process, limits their achievable speed-up. To address these challenges, dedicated caching strategies~\cite{liu2025dllm,ma2025dkv,wu2025fast} have been developed to reuse computations and approximate bidirectional attention. In addition, to realize the potential of parallel token generation, confidence-based sampling~\cite{wu2025fast}, guidance from AR models~\cite{israel2025accelerating}, and adaptive decoding with certainty and positional priors~\cite{wei2025accelerating} have been proposed. 
Beyond these training-free methods, \cite{gong2025scaling,ye2025dream} propose initializing diffusion LMs from AR models with token shifts to accelerate diffusion LM training. Block Diffusion~\cite{arriola2025block} combines AR and diffusion by performing block-wise AR and in-block diffusion to support native KV caching. Follow-up works~\cite{wu2025fast2,fu2025efficient,cheng2025sdar,wang2025diffusion,qian2026d3llm} also convert pretrained AR models or diffusion LMs into block-wise ones.~\cite{gat2025set,samragh2025your} further explore combining AR and diffusion through either joint training or LoRA modules dedicated to diffusion.

\section{Insights and Future Directions}
\label{sec:conclusion}

We deliver \METHOD{}, a tri-mode language model family trained via joint AR-diffusion optimization that unifies AR, diffusion, and self-speculation within a single model. The resulting base, instruct, and vision-language models outperform SOTA open-source AR/diffusion LMs in both accuracy and efficiency. The training and analysis of tri-mode LMs reveal the following insights:

\begin{enumerate}[leftmargin=*,itemsep=2pt]

\item \textbf{Tri-mode generation arises naturally from joint AR-diffusion training.} By enabling both AR and non-AR parallel token prediction within a single model, the joint training objective simultaneously produces three inference modes without any mode-specific architectural modifications.

\item \textbf{AR and diffusion losses are complementary, not competing.} The two objectives mutually benefit each other and peak at the same loss coefficient ($\alpha{=}0.3$). Adding the AR loss induces left-to-right linguistic priors for diffusion, and the diffusion loss preserves or slightly improves AR accuracy through better future planning.

\item \textbf{Self-speculation outperforms MTP methods.} Instead of relying on auxiliary prediction heads, self-speculation leverages diffusion to generate high-quality multi-token drafts and uses AR verification to ensure correctness, achieving higher acceptance rates and better efficiency.

\item \textbf{Variance reduction is critical for diffusion training.} The diffusion loss introduces intrinsically high variance due to random masking with variable noise levels. More sufficiently trained AR starting points (e.g., via two-stage training) or variance-reduction training techniques (e.g., global loss averaging) can improve training effectiveness.

\item \textbf{Linear self-speculation is currently the most efficient mode.} Linear self-speculation achieves the best efficiency in the current infrastructure. Quadratic self-speculation achieves higher TPF per step, making it more promising at batch size 1 with improved infrastructure support.

\item \textbf{Diffusion-mode decoding has substantial headroom.} Our SOL analysis shows the potential to correctly predict 76.5\% more tokens per forward pass than the current best strategy (linear self-speculation), indicating a more promising upper bound for parallel decoding than speculative decoding based only on prefix decoding.

\end{enumerate}

Looking forward, these insights shed light on several promising directions for further improvements:

\begin{enumerate}[leftmargin=*,itemsep=2pt]

\item \textbf{Closing the gap between practical diffusion decoding and its SOL upper bound.} Our SOL analysis suggests that diffusion-mode decoding could offer a more attractive path toward parallel decoding than linear decoding, due to its non-prefix acceptance pattern and therefore higher upper bound. However, current confidence-based samplers remain far from this upper bound. Developing optimized samplers that more reliably identify correct tokens, or more advanced training schemes that enable more aggressive parallel sampling of conditionally independent tokens, is a promising direction for closing this gap.

\item \textbf{Improving draft-verification alignment for self-speculation.} Given the strong practical speedup achieved by self-speculation, an important future direction is to better align the diffusion draft mode with the AR verification mode during training, thereby improving the acceptance rate. In addition, the cost of drafting can be further reduced by using nested subnets, where weight-shared smaller subnets generate drafts through specialized training techniques~\cite{cai2024flextron,fu2024amoeballm}.

\item \textbf{Beyond prefix-only AR verification.} Current AR verification accepts drafted tokens only in a prefix-wise manner, which does not fully exploit the non-AR nature of diffusion drafts. A promising direction is to explore diffusion-mode verification, potentially using another diffusion verifier, to validate multiple non-contiguous drafted tokens and further improve the effective acceptance rate.

\item \textbf{Enabling higher-level parallelism in diffusion generation.} Although diffusion decoding enables parallel token prediction, its generation order still exhibits a strong left-to-right tendency and mainly provides token-level parallelism. Future training algorithms that encourage segment-level or paragraph-level parallelism could better unlock the global planning ability and efficiency potential of diffusion-mode generation.

\end{enumerate}

{
  \small
  \bibliographystyle{unsrt}
  \bibliography{ref}
}

\appendix

\newpage

\section{Diffusion Sampler Details}
\label{sec:appendix_sampler}

This appendix details the sampler introduced in Sec.~\ref{sec:diff_mode_inference}, including its architecture, input features, and training trajectory collection.

\begin{figure}[b]
\centering
\vspace{-1em}
\includegraphics[width=\linewidth]{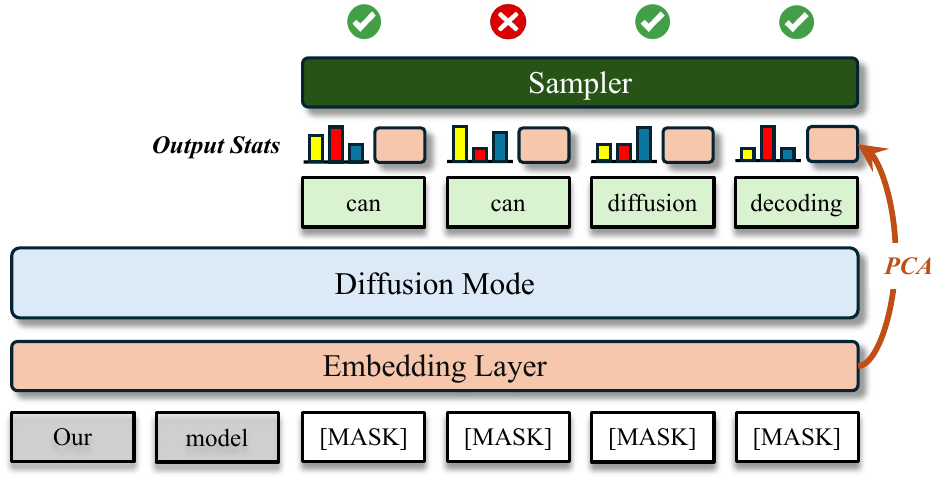} 
\vspace{-1.5em}
\caption{An illustration of the sampler design on top of the diffusion mode.
}
\label{fig:sampler_design}
\end{figure}

\noindent \underline{Architecture and feature engineering.} As shown in Fig.~\ref{fig:sampler_design},
the sampler operates on top of the frozen backbone and adds negligible parameter overhead ($\sim$$0.06\%$, with $4.8\mathrm{M}$ compared to the $8\mathrm{B}$ backbone).
It is a 4-layer lightweight Transformer with a hidden dimension of $d{=}384$. It attends bidirectionally over the current block, followed by a per-position linear head with a sigmoid output.
Each input position is represented by a 144-dimensional feature: PCA-compressed semantic embeddings of the top-$3$ predictions, as well as statistics summarizing the output distribution (e.g., top-$1$ probability, margin, top-$3$ mass, and entropy). 
The semantic embedding of the model’s own top-$1$ prediction is by far the most informative feature. We also find cross-position attention to be essential as an MLP-only ablation drops accuracy-TPF AUC by 10 percentage points, indicating that the sampler must jointly reason about which positions are mutually safe to commit.

\noindent \underline{Data collection for sampler training.}
To train the sampler, we collect $\sim$$20\mathrm{M}$ denoising trajectories from \METHOD{}-8B on~\cite{NemotronPostTrainingDatasetV2} (math, code, STEM, and chat subsets) at block lengths $B \in {8, 32}$. We use two complementary trajectory policies: (i) standard confidence decoding ($k{=}1$), where the block is decoded one token at a time in confidence order; and (ii) a hybrid policy that first commits ground-truth tokens whenever the model’s top-$1$ prediction already agrees with them, then falls back to confidence for the remaining positions.
At each intermediate step of every trajectory, we store the 144-dimensional per-position features and the binary label $\mathbf{1}[\text{current top-}1 = \text{final ID}]$, where the \emph{final ID} is the token ultimately committed at that position once the block is fully decoded under the same policy. Training uses per-position binary cross-entropy on masked positions, with AUC on a held-out trajectory split as the early stopping criterion.
The resulting accuracy–TPF gains over confidence thresholding are empirically reported in Sec.~\ref{sec:exp_instruct} and Fig.~\ref{fig:sampler}.

\section{LoRA-Enhanced Linear SS}
\label{sec:appendix_linear_ss}

\begin{figure*}[t]
\centering
\vspace{-0.5em}
\includegraphics[width=0.99\linewidth]{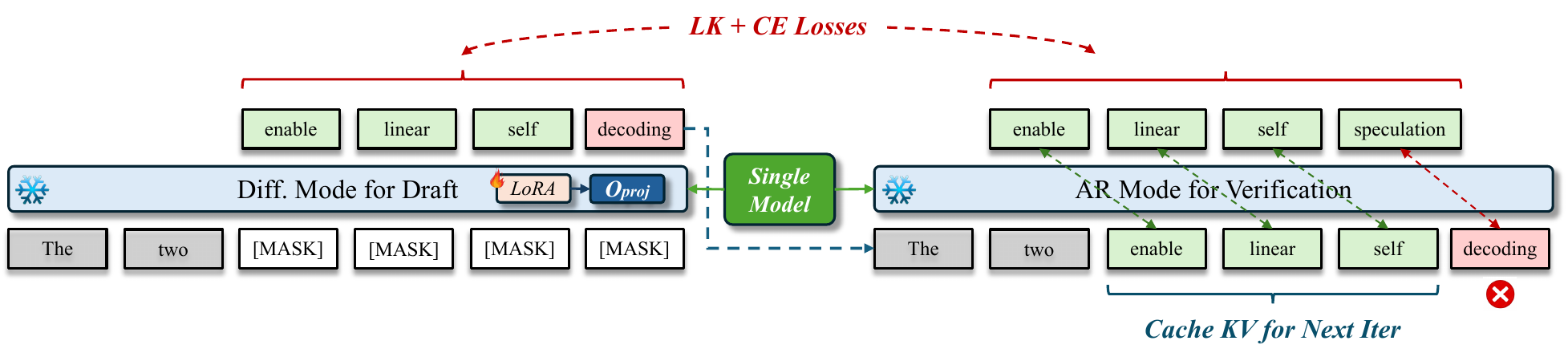} 
\vspace{-1em}
\caption{
An illustration of LoRA training on the diffusion drafter of the linear self-speculation mode.
}
\label{fig:lora_loss}
\vspace{-1em}
\end{figure*}

We provide a visualization of the enhanced linear self-speculation w/ LoRA in Fig.~\ref{fig:lora_loss}.

\section{Quadratic SS Details}
\label{sec:appendix_quad_ss}

\begin{figure}[t]
\centering
\includegraphics[width=0.99\linewidth]{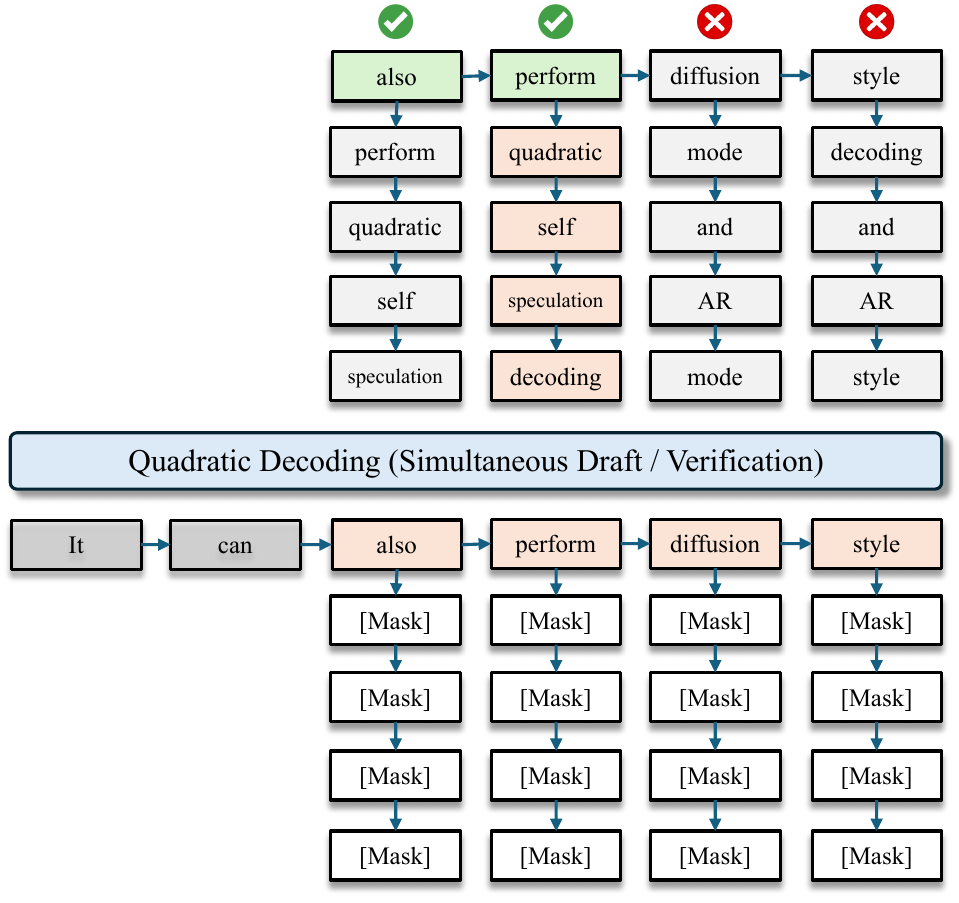} 
\vspace{-1.5em}
\caption{
An illustration of quadratic self-speculation with simultaneous drafting and verification.
\tinybox{maskgreen} denotes draft tokens that match AR verification, and \tinybox{maskorange} denotes the block that originates from the last matched token and is to be verified in the next iteration.
}
\label{fig:quadratic_speculation}
\vspace{-1.5em}
\end{figure}

We provide more details about quadratic self-speculation, which is visualized in Fig.~\ref{fig:quadratic_speculation}.
Specifically, let $[x_1,\ldots,x_n]$ denote the currently verified prefix, and let $k$ be the speculative width. At generation step $t{+}1$, we reuse the $k$ speculative tokens from the previous step, denoted $\{x^{t}_{n+j}\}_{j=2}^{k+1}$, and interleave $k$ fresh mask tokens after each speculative token, yielding the quadratic input:
\begin{equation}
\begin{aligned}
X^{t+1}_m
&=
[x_1,\ldots,x_n, x_{n+1}]
\;+\;
[x^{t}_{n+2}, m_1,\ldots,m_k] \\
&\quad+\;\cdots+\;
[x^{t}_{n+k+1}, m_1,\ldots,m_k],
\end{aligned}
\label{eq:quad_input}
\end{equation}
where $x_{n+1}$ is the next token generated autoregressively at step $t{+}1$ (and is thus immediately verified), and the total number of inserted masks is $k^2$.

\underline{Parallel draft and verification.}
Feeding $X^{t+1}_m$ via a structured attention mask into our model produces two types of outputs in a single forward pass.
First, the model generates next-token predictions for the speculative tokens in a causal manner, yielding $\{x^{t+1}_{n+j}\}_{j=2}^{k+1}$. These tokens are used to verify the previous speculative draft through sequential comparison~\cite{samragh2025your}: we accept the longest prefix that satisfies a verification criterion (e.g., $x^{t+1}_{n+j}=x^{t}_{n+j}$, as detailed later), commit the accepted tokens to the verified prefix, and stop verification at the first mismatch.
Second, in the same forward pass, the model predicts the tokens corresponding to the newly inserted masks $\{m_r\}_{r=1}^{k}$ in parallel; we treat these predictions as the draft tokens that will serve as $\{x^{t+1}_{n+j}\}_{j=2}^{k+1}$ in the next iteration.
The interleaved quadratic layout ensures that, even if verification fails early at some position, newly inserted mask positions remain that still yield fresh speculative tokens for the next step, so each iteration consistently produces $k$ tokens to verify~\cite{samragh2025your}. 

\underline{Verification with the AR-diffusion ensemble.}
For verification, the simplest choice is to use the AR predictions on the speculative tokens. In addition, our tri-mode model provides a complementary verification signal from the diffusion pathway: for each speculative token $x^{t}_{n+j}$ in Eq.~\ref{eq:quad_input}, we can use the diffusion prediction at the first newly inserted mask position $m_1$ immediately following it as an alternative verifier for the same token. Concretely, the AR verifier uses the causal logits $p_{\theta}^{\text{AR}}(\cdot \mid x_{<n+j})$ at position $n{+}j$, while the diffusion verifier uses the denoising logits $p_{\theta}^{\text{diff}}(\cdot \mid X^{t+1}_m, t)$ produced for the corresponding $m_1$ position. 
Generally, we can form an AR-diffusion ensemble verifier by combining the two distributions:
\[
p^{\text{ens}}_{\theta}(\cdot) = \lambda\, p^{\text{AR}}_{\theta}(\cdot) + (1-\lambda)\, p^{\text{diff}}_{\theta}(\cdot),
\]
where $\lambda \in [0,1]$ controls the interpolation.

\end{document}